\documentclass{article}
\usepackage[T1]{fontenc}
\usepackage{iclr2027_conference,times}
\usepackage{algorithm}
\usepackage{algpseudocode}
\usepackage{wrapfig}
\newsavebox{\codetabbox}   
\usepackage{capt-of}       
\usepackage{caption}
\usepackage{subcaption}
\usepackage{listings}
\usepackage{titletoc}
\usepackage{amsfonts}
\iclrfinalcopy

\usepackage[most]{tcolorbox}

\newtcblisting{locdprompt}[1]{
    enhanced,
    breakable,
    listing only,
    title={#1},
    colback=black!1,
    colframe=black!35,
    colbacktitle=black!6,
    coltitle=black,
    fonttitle=\bfseries\small,
    boxrule=0.4pt,
    arc=1.5pt,
    boxsep=1.5mm,
    left=1mm,
    right=1mm,
    before skip=8pt,
    after skip=8pt,
    listing options={
        basicstyle=\ttfamily\footnotesize,
        breaklines=true,
        columns=fullflexible,
        keepspaces=true,
        showstringspaces=false
    }
}

\usepackage{hyperref,url,graphicx,amsmath,booktabs,colortbl,xcolor,makecell}
\usepackage[section]{placeins}
\definecolor{bestbg}{RGB}{249,231,178}
\definecolor{secondbg}{RGB}{221,233,245}
\newcommand{\best}[1]{\cellcolor{bestbg}\textbf{#1}}
\newcommand{\second}[1]{\cellcolor{secondbg}\underline{#1}}
\newcommand{\std}[1]{_{\scriptstyle\,\pm\,#1}}

\title{Learning from Interaction Experience via\\Localized Context Distillation}
\title{Learning from Interaction Experience via \\
Localized Context Distillation}

\author{%
\bfseries
Zhezheng Hao\textsuperscript{1\,*},
Yongkang Yang\textsuperscript{2\,*},
Xilin Xia\textsuperscript{2\,*},
Jiarui Yu\textsuperscript{2},
Yi Liu\textsuperscript{3},
Fanjunduo Wei\textsuperscript{1},\\
\bfseries
Peijie Dong\textsuperscript{4},
Tianfu Wang\textsuperscript{4},
Qiang Lin\textsuperscript{2},
Can Wang\textsuperscript{1},
Hande Dong\textsuperscript{2\,\textdaggerdbl},
Jiawei Chen\textsuperscript{1\,\textdagger\,\textdaggerdbl}
\\[1.0ex]
\mdseries
\textsuperscript{1}Zhejiang University
\,\,
\textsuperscript{2}Workbuddy Team, Tencent
\,\,
\textsuperscript{3}The Chinese University of Hong Kong
\\
\textsuperscript{4}The Hong Kong University of Science and Technology
}

\begin{document}
\maketitle

\begingroup
\renewcommand{\thefootnote}{\fnsymbol{footnote}}
\footnotetext[1]{Equal contribution.\quad \textsuperscript{\textdagger}Corresponding author.\quad \textsuperscript{\textdaggerdbl}Project leader.}
\endgroup
\begin{abstract}
Learning from interaction experience is essential for enabling large language models to continually improve during deployment. Context distillation offers a promising path: privileged information extracted from experience is provided to a teacher, whose behavior is then distilled into a student model that has no access to it. However, both existing forms of context distillation have limitations. Off-policy context distillation trains the student on teacher-generated trajectories and thus suffers from exposure bias; whereas on-policy context distillation trains on the student's own rollouts, whose states are often uninformative and fail to convey behaviors the student cannot yet produce. To resolve this dilemma, we introduce \textbf{Localized Context Distillation (LoCD)}. LoCD first localizes the critical spans of a student trajectory that require revision. The teacher then regenerates only these spans from the student's own prefix, and the student is trained on the regenerated segments. LoCD thereby concentrates learning on the most informative states while keeping exposure bias low. Extensive experiments across math reasoning, code generation, and agent tasks demonstrate that LoCD consistently outperforms strong baselines and achieves more stable training with smaller and sparser parameter updates.\footnotetext[3]{Code is available at \url{https://github.com/zz-haooo/LoCD}.}
\end{abstract}

\vspace{-1mm}
\section{Introduction}
\label{sec:introduction}
\vspace{-1mm}
Large language models (LLMs) acquire most of their capabilities before deployment, through supervised fine-tuning on curated demonstrations and reinforcement learning with verifiable rewards~\citep{ouyang2022training,guo2025deepseek}.
Once deployed, however, a model inevitably meets situations its training never covered and makes mistakes, creating a persistent need for further learning.
Fortunately, each such mistake leaves behind an \textbf{interaction experience}, i.e., the model's own trajectory together with feedback from the environment or the user, which reveals where the behavior went wrong and how it can be revised~\citep{chen2025retrospective,wang2026text2grad,zhang2025critique}.
This raises a central research question: \textbf{how can LLMs learn from interaction experience so that they continually improve during deployment?}

To tackle this, a broad line of work stores experience in external memory and retrieves it as additional context when similar situations recur~\citep{zhao2024expel,chhikara2025mem0,hao2026recreate}.
However, as interaction experience continues to accumulate, retrieving relevant experience becomes increasingly difficult, and noisy or mismatched experience may interfere with model behavior~\citep{salama2025meminsight,chao2026stale}.
In contrast, \emph{Context Distillation} offers a parametric alternative that internalizes useful knowledge from experience directly into model parameters: the experience-derived context is treated as \textbf{privileged context}, available to a teacher during training but absent at inference; the same model without the context acts as the student and is trained to match the teacher's behavior~\citep{askell2021general,snell2022learning,hu2024teaching}.
Existing context distillation methods mainly fall into two categories.

Off-policy context distillation allows the privileged teacher to generate a revised trajectory and trains the student to reproduce it~\citep{scheurer2023training,bai2022constitutional}.
This provides direct supervision on the revised behavior, but it creates a mismatch between training and inference known as \emph{exposure bias}: the student is only ever supervised on states visited by the teacher, yet at inference it must condition on states produced by its own policy.
Such exposure bias can make it difficult to reproduce teacher behavior over long trajectories and may further exacerbate catastrophic
forgetting~\citep{wang2026backtracking,shenfeld2026self}.

On-policy context distillation addresses this mismatch by training on the student's trajectories instead~\citep{zhao2026self,shenfeld2026self,ye2026policy,wang2026openclaw}: the student samples a trajectory on its own, and the privileged teacher provides token-level supervision at each student-generated prefix.
Training and inference thus share the same state distribution. However, we find that restricting supervision entirely to student-generated trajectories introduces another bottleneck: on-policy context distillation readily reinforces behaviors that already appear in the student's own rollouts, but struggles to directly internalize new behaviors induced by privileged context.
Empirically, as the student's initial accuracy decreases, on-policy context distillation becomes substantially less effective.
On cases where the student fails across all sampled rollouts, off-policy distillation still learns substantially from the privileged context, whereas on-policy context distillation yields only marginal improvement.
To probe the learning mechanism behind this limitation, we further conduct a controlled knowledge-injection experiment in which we vary the length of the answer to be learned.
We observe a progressive learning pattern: longer answers require progressively more training to learn.
This pattern is consistent with a sequential learning bottleneck: supervision for later parts becomes more informative as prefixes compatible with the target behavior appear more frequently in student rollouts.
Such bottleneck of on-policy context distillation makes behaviors that deviate from the student's current distribution difficult to learn.

Given the weaknesses of the two paradigms, we propose a new strategy, called \textbf{Localized Context Distillation (LoCD)}.
Its core idea is to introduce a short segment of teacher-generated behavior only where the student-generated trajectory requires revision.
Specifically, LoCD first localizes a \emph{critical span} of the student trajectory that requires revision and attributes the corresponding privileged context to that span.
It then performs localized distillation:  starting from the student's own prefix, the privileged teacher regenerates only this span, and the student is trained on the teacher-generated segment.
This design places supervision on the most useful states: the span starts at the critical point where the student trajectory begins to go wrong, and the states within it are generated by the teacher, so the supervision is informative.
At the same time, the teacher continues from the student's own prefix and intervenes only within this short span, so the training states stay close to the student's distribution and exposure bias remains low.


Extensive experiments across math reasoning, code generation,
interactive decision-making, and software-engineering tasks show that LoCD
consistently outperforms strong baselines for learning from interaction experience.
Compared with the strongest baselines, LoCD improves performance by 5.65\% on math reasoning, 2.65\% on code generation, 5.60\% on WebShop, and 7.46\% on ALFWorld.
Beyond final performance, LoCD exhibits more stable training dynamics and avoids
the sustained policy-entropy collapse observed in on-policy and off-policy
context distillation.
Further analysis shows that these gains are achieved with substantially smaller
and sparser parameter updates, supporting the effectiveness of restricting
supervision to critical spans.

\vspace{-1mm}
\section{Preliminaries}
\label{sec:preliminaries}
\vspace{-1mm}
\subsection{Interaction Experience and Privileged Context}

We consider a language model that is given a task $x$ and produces an
interaction trajectory
\[
\tau=(a_1,f_1,\ldots,a_K,f_K),
\]
where $a_k$ denotes the model output at step $k$, such as a reasoning step, an
action, or a response, and $f_k$ denotes the feedback provided after that output.
We refer to $(x,\tau)$ as an \emph{interaction experience}.
In interactive environments, the feedback $f_k$ can take the form of execution
results, tool responses, or environment observations.
For single-step reasoning and coding tasks without native interaction feedback,
we construct an analogous interaction experience by using a reference solution
as feedback for the model output, following prior work
~\citep{kara2026role,zhao2026rosd,wang2026hindsight}.

The feedback contained in an interaction experience can provide information about errors in the model's reasoning or actions and how its behavior should be improved~\citep{scheurer2023training,hao2026recreate,wang2026envrl}.
Following prior work~\citep{ye2026online,liu2026hero}, we use the model itself to extract a compact context from each interaction experience:
\[
c \sim \mathcal{E}(\cdot\mid x,\tau),
\]
where $\mathcal{E}$ denotes the experience extractor and $c$ is the resulting \emph{privileged context}.
We study how to internalize the useful information in the privileged context $c$
into model parameters, so that the model can improve future behavior without
accessing the original interaction experience at inference time.

\subsection{Context Distillation}

Context distillation provides a framework for learning from interaction experience
by using a teacher with access to privileged context to supervise a student without
access to that context~\citep{askell2021general,snell2022learning}.
In our setting, the experience-derived context $c$ serves as the privileged
context for distillation~\citep{hubotter2026reinforcement,liu2026hero,feng2026past}.

Let $y=(y_1,\ldots,y_T)$ denote the trajectory used for training.
We denote the student and privileged teacher by
$\pi_S(\cdot\mid x,y_{<t})$ and
$\pi_T(\cdot\mid x,c,y_{<t})$, respectively.
In the self-distillation setting studied in this work, they share the same
underlying model and differ only in whether the privileged context $c$ is
provided.
Formally, given a policy $\pi$, context distillation minimizes the token-level divergence between
the privileged teacher and the student,
\[
\mathcal{L}_{\mathrm{CD}}
=
\mathbb{E}_{
(x,\tau)\sim\mathcal{D},\,
c\sim\mathcal{E}(\cdot\mid x,\tau),\,
y\sim\pi
}
\left[
\frac{1}{|y|}
\sum_{t=1}^{|y|}
D\!\left(
\operatorname{sg}\!\left[
\pi_T(\cdot\mid x,c,y_{<t})
\right],
\pi_S(\cdot\mid x,y_{<t})
\right)
\right].
\]
Here, $\mathcal{D}$ denotes the set of interaction experiences
collected online from the current student, $\operatorname{sg}[\cdot]$ denotes stop-gradient, and $D$ denotes a
distributional divergence, such as forward KL, reverse KL, or
Jensen--Shannon divergence
~\citep{agarwal2024policy,zhao2026self}.
Off-policy and on-policy context distillation differ in the rollout policy
$\pi$ used to generate the distillation trajectory $y$.
For off-policy context distillation, $\pi=\pi_T$ and
$y\sim\pi_T(\cdot\mid x,c)$, whereas for on-policy context distillation,
$\pi=\pi_S$ and $y\sim\pi_S(\cdot\mid x)$.

\vspace{-1mm}

\section{Empirical Study}
\label{Empirical Study}
\vspace{-1mm}

Recent studies find that on-policy context distillation does not consistently improve performance across a range of settings
~\citep{zhu2026many,shen2026purified,kaur2026rethinking}.
This leaves open whether on-policy context distillation can effectively learn behaviors induced by privileged context, especially when those behaviors are rarely or never produced by the student itself.
We therefore examine how its ability to learn such behaviors varies with the student's initial accuracy, using accuracy as a measure of how frequently successful behavior appears in the student's own rollouts.
We further analyze the role of privileged context in
Appendix~\ref{app:pi_semantics}.

\paragraph{Experimental setup.}
We conduct this empirical study on the widely-used math reasoning dataset  \textsc{OpenThoughts}~\citep{guha2026openthoughts} using three models:
Qwen3-4B, Qwen3-8B, and Falcon3-10B-Instruct.
We construct the interaction experiences and privileged context following
Section~\ref{sec:preliminaries}, with full details provided in
Appendix~\ref{Details of the Initial-Accuracy Study}.

For each problem, we measure the student's initial accuracy over 128 independent
rollouts (avg@128) and partition problems into bins according to this accuracy.
We sample 100 problems from each bin and train both methods for 5 epochs,
using OPCD~\citep{ye2026policy} for on-policy context distillation and
TRD~\citep{jiang2026trajectory} for off-policy context distillation.
After training, we evaluate the distilled student without privileged context
on the same problems used for training.

Let $P_{\mathrm{base}}$ and $P_T$ denote the performance of the initial student
and the privileged teacher, respectively.
We define the additional performance enabled by privileged context as the
\emph{privileged improvement},
$\Delta_{\mathrm{priv}} = P_T-P_{\mathrm{base}}$.
We compare the distilled student's improvement over $P_{\mathrm{base}}$ with
$\Delta_{\mathrm{priv}}$ to measure how effectively it learns from the
privileged context.

\begin{figure}[t]
    \centering
    \includegraphics[width=\linewidth]{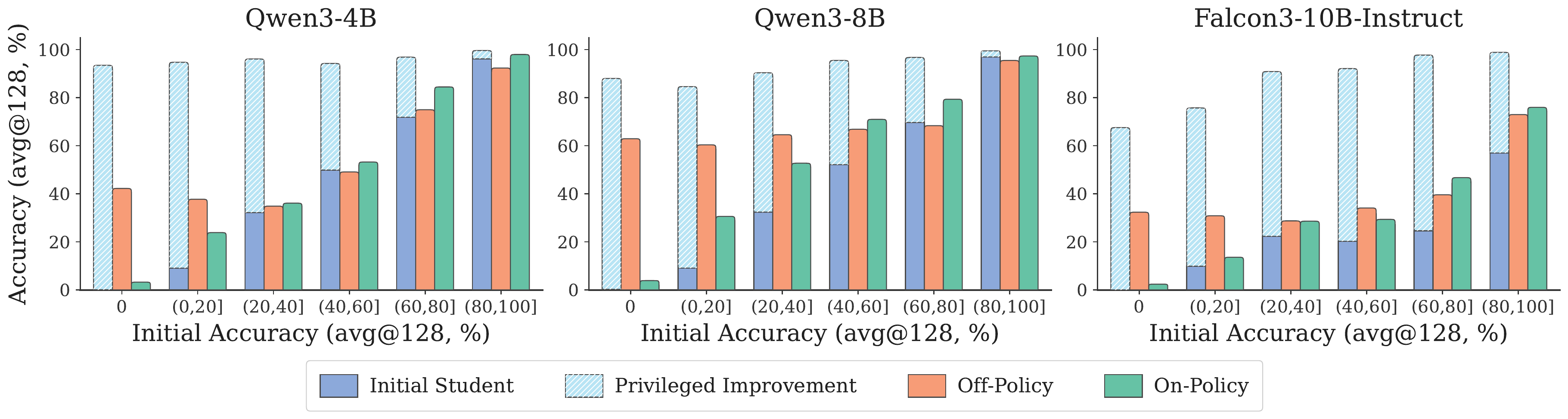}
    \vspace{-5mm}
    \caption{How initial student accuracy affects learning from privileged context.}
    \label{How Initial Student Accuracy Affects Learning from Privileged Context.}
    \vspace{-1mm}
\end{figure}


\paragraph{On-policy context distillation becomes less effective at low initial accuracy.}
As shown in Figure~\ref{How Initial Student Accuracy Affects Learning from Privileged Context.}, all three models exhibit a
consistent pattern: the effectiveness of on-policy context distillation decreases
as the student's initial accuracy becomes lower.
When the student's initial accuracy is above $60\%$, on-policy context distillation retains a substantial portion of the privileged improvement and outperforms off-policy context distillation.
At lower initial accuracy levels, however, this advantage disappears, and
on-policy context distillation becomes substantially less effective than
off-policy context distillation.

The gap between the two methods is most pronounced for problems with zero
initial accuracy (avg@128 $=0$).
Although the initial student produces no correct output among 128 independent
rollouts, the privileged teacher still substantially outperforms the initial student,
showing that the privileged context provides useful information for these problems.
This provides a clean setting for testing whether the two distillation methods can learn from useful privileged context when the desired behavior is absent from the student's own rollouts.
As shown, off-policy context distillation substantially improves the student, whereas
on-policy context distillation yields only marginal improvement.

\begin{figure}[t]
    \centering
    \includegraphics[width=\linewidth]{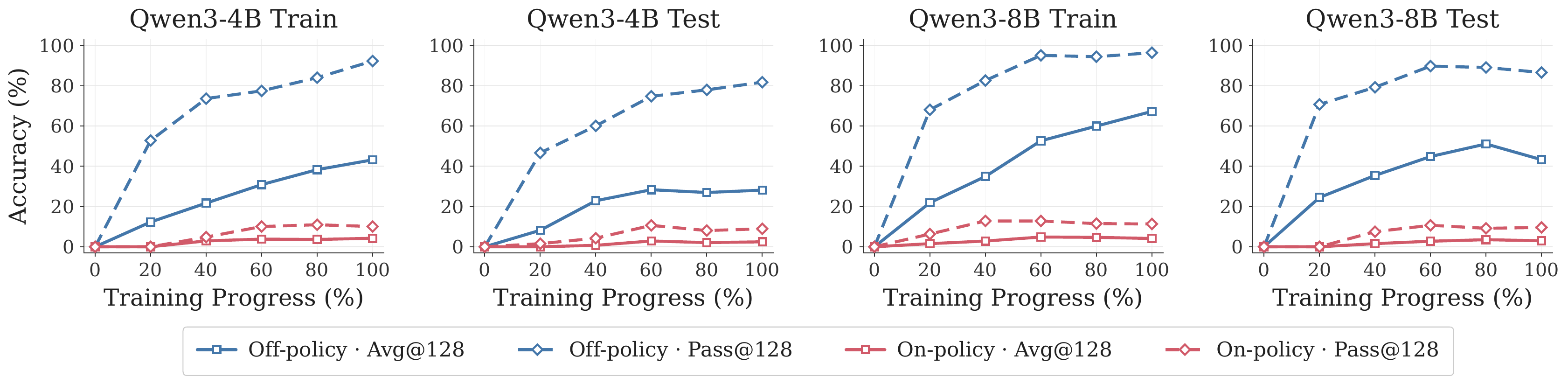}
    \vspace{-5mm}
    \caption{Training dynamics on problems with zero initial accuracy (avg@128 = 0).}
    \label{initial accuracyTraining dynamics on problems with zero initial accuracy (avg@128 = 0).}
    \vspace{-1mm}
\end{figure}

We further examine whether this gap persists with continued training.
Figure~\ref{initial accuracyTraining dynamics on problems with zero initial accuracy (avg@128 = 0).} tracks avg@128 and pass@128 throughout
training on problems with zero initial accuracy.
For both models, off-policy context distillation improves rapidly early in training and continues to achieve substantially higher avg@128 and pass@128.
In contrast, on-policy context distillation shows only limited improvement, and the gap between the two methods persists throughout training.
To examine whether this pattern extends beyond the training problems, we construct a held-out set of 100 problems with zero initial accuracy.
The same dynamics emerge on the held-out set: off-policy context distillation improves substantially, whereas on-policy context distillation remains far below it throughout training.

Across these experiments, on-policy context distillation becomes less effective
when behaviors induced by privileged context rarely appear in the student's
own rollouts.
This phenomenon points to a simple and important mechanism: on-policy context distillation applies
teacher supervision only on student-generated prefixes, which creates a sequential dependency:
later parts of a new behavior become effectively learnable only after the preceding parts begin to appear in the student's own rollouts.
Consequently, longer target behaviors should require progressively more training to learn.

\paragraph{On-policy context distillation exhibits length-dependent learning.}
To further examine this explanation, we construct a controlled
knowledge-injection task following prior work
~\citep{chang2024large,zhang2026bidirectional}.
Each example consists of a synthetic fact and a corresponding question, such as
``\textit{Sne's coach is Cour}'' and
``\textit{What is Sne's coach?}''
The synthetic fact is directly used as privileged context during training.

\begin{wrapfigure}{r}{0.38\textwidth}
    \centering
    \vspace{-3mm}
    \includegraphics[width=\linewidth]{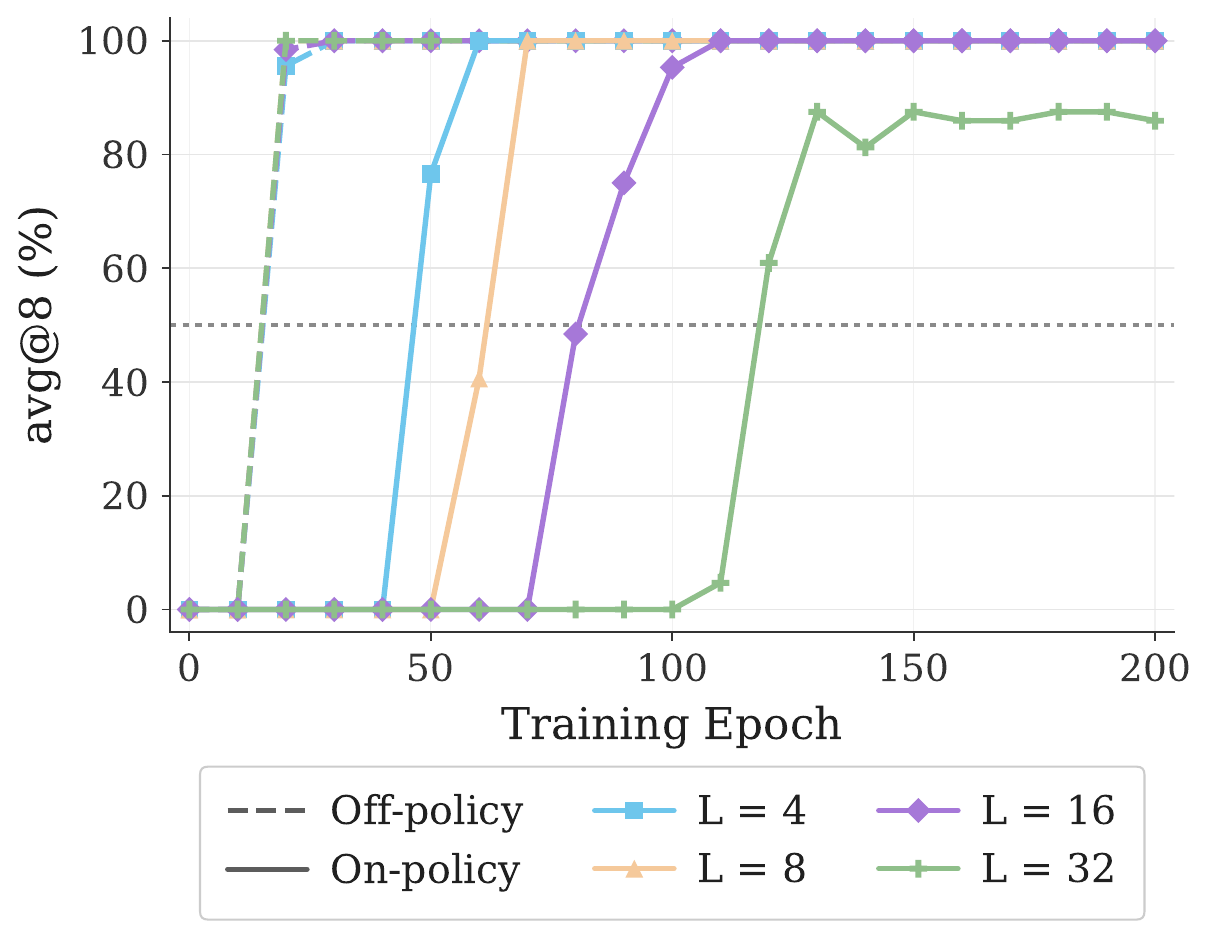}
    \vspace{-6mm}
    \caption{Knowledge-injection dynamics across answer lengths.}
    \label{Knowledge-injection dynamics across different answer lengths.}
    \vspace{-8mm}
\end{wrapfigure}

We vary the length of the answer to be learned, with
$L\in\{4,8,16,32\}$ tokens, and mix all length groups within the same training
run.
We use the same on-policy and off-policy context distillation methods as above,
training for 200 epochs and evaluating every 5 epochs.
Full experimental details are provided in
Appendix~\ref{app:knowledge_injection}.

As shown in Figure~\ref{Knowledge-injection dynamics across different answer lengths.}, off-policy context distillation
rapidly learns answers of all lengths, with little sensitivity to answer length.
In contrast, on-policy context distillation exhibits a clear length-dependent
delay: performance begins to improve earliest for $L=4$, followed by
$L=8$, $L=16$, and $L=32$.
Moreover, the $L=32$ group remains below the shorter-length groups under the
same training budget.

This length-dependent delay supports the sequential dependency described above.
Although the complete answer is available to the privileged teacher from the
beginning, later parts become effectively learnable only after the preceding
parts begin to appear in the student's own rollouts.

These findings are particularly relevant to learning from failed interaction
trajectories.
Such trajectories arise precisely when the current policy does not yet produce
the desired behavior, making the required correction less likely to appear in
the student's own rollouts.
This places learning from failed interactions in the regime where on-policy
context distillation is less effective.



\vspace{-1mm}

\section{Localized Context Distillation}
\label{sec:method}
\vspace{-1mm}
Our empirical study shows that on-policy context distillation becomes much less
effective when the behaviors induced by privileged context rarely appear in the
student's own rollouts.
The key limitation is that privileged supervision depends on student-generated
prefixes, making behaviors that the current policy cannot reliably produce
difficult to incorporate into training.

A natural alternative is to let the privileged teacher directly generate a
revised trajectory, as in off-policy context distillation.
While this can explicitly introduce behaviors absent from the student's own
rollouts, training on teacher-generated states introduces exposure bias, which
may further contribute to catastrophic forgetting or reasoning collapse
~\citep{shenfeld2026self,kim2026does}.

\begin{figure}[t]
    \centering
    \includegraphics[width=0.9\linewidth]{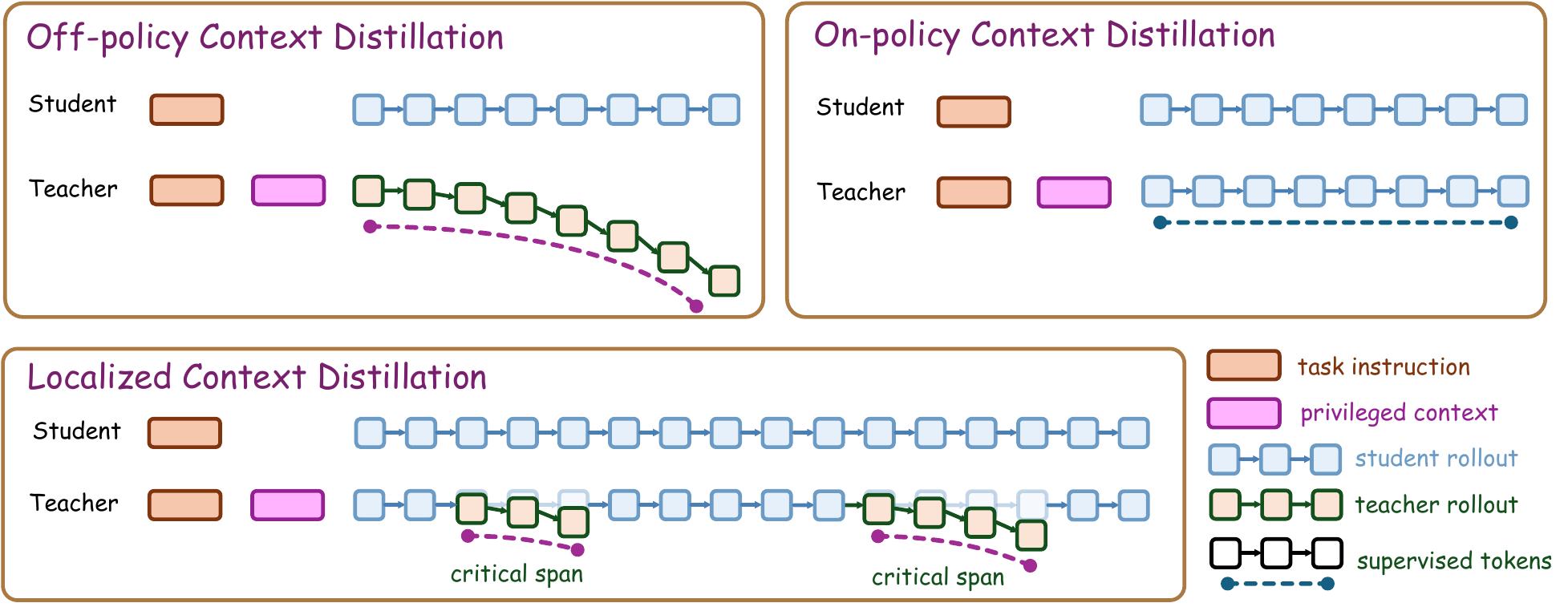}
    \vspace{-1mm}
    \caption{Illustrative comparison of off-policy, on-policy, and localized context distillation.}
    \label{Illustrative comparison of off-policy, on-policy, and localized context distillation.}
    \vspace{-1mm}
\end{figure}

To address the limitations of both approaches, we introduce
\emph{Localized Context Distillation} (LoCD).
As illustrated in Figure~\ref{Illustrative comparison of off-policy, on-policy, and localized context distillation.}, LoCD preserves the
student-generated prefix and introduces privileged teacher generation only
within an attributed \emph{critical span}.
Specifically, LoCD consists of two steps:
\emph{(i) localization and attribution} and \emph{(ii) localized distillation}.
After each update, the resulting student policy is used to collect the next
round of interaction experience.

\paragraph{Localization and attribution.}
Given an interaction experience $(x,\tau)$, LoCD localizes the spans that
require revision and extracts their corresponding privileged contexts:
\[
([\alpha,\beta),c)\sim\mathcal{A}(\cdot\mid x,\tau).
\]
Here, $\mathcal{A}$ is a prompt-based localization and attribution procedure
implemented by prompting the same model as the student, leveraging its own reasoning
capabilities.
The indices $\alpha$ and $\beta$ denote the start and end indices of an attributed
\emph{critical span} along the student trajectory, and $c$ denotes its corresponding privileged context.
For example, a critical span can correspond to a contiguous reasoning segment or a specific interaction step in an agent trajectory.
When multiple critical spans are identified from the same interaction, we treat each attributed span as a separate training instance.

\paragraph{Localized distillation.}
Given an attributed critical span $([\alpha,\beta),c)$, LoCD samples a local sequence
$y^T$ from the privileged teacher starting from the student-generated prefix
$y^S_{<\alpha}$.
We optimize
\[
\mathcal{L}
=
\mathbb{E}_{\substack{
(x,\tau)\sim\mathcal{D},\;
([\alpha,\beta),c)\sim\mathcal{A}(\cdot\mid x,\tau)
\\
y^T\sim\pi_T(\cdot\mid x,c,y^S_{<\alpha})
}}
\left[
\frac{1}{\beta\!-\!\alpha}
\sum_{t=1}^{\beta-\alpha}
D\!\left(
\operatorname{sg}\!\left[
\pi_T(\cdot\mid x,c,y^S_{<\alpha},y^T_{<t})
\right],
\pi_S(\cdot\mid x,y^S_{<\alpha},y^T_{<t})
\right)
\right].
\]
Here, teacher sampling is truncated after position $\beta$, restricting privileged
generation to the attributed critical span.
The divergence $D$ can be forward KL,
reverse KL, or Jensen--Shannon divergence
~\citep{agarwal2024policy,zhao2026self}.
Following prior work~\citep{zhao2026self,yang2026matching},
we clip excessively large token-level divergences
to stabilize training.

Unlike off-policy context distillation, which supervises
teacher-generated trajectories, and on-policy context
distillation, which supervises student-generated prefixes,
LoCD restricts privileged supervision to localized
teacher-generated segments.
This design offers several advantages as follows.

\paragraph{Advantages of LoCD.}
First, LoCD anchors each correction at a state actually visited by the current
student, while allowing the privileged teacher to directly supervise behaviors
that the student cannot yet produce.
This enables corrective supervision directly where the student requires revision.
Second, by restricting teacher-generated prefixes to the critical span, LoCD
confines teacher-induced deviation to the region that requires correction,
thereby limiting exposure bias.
Third, through localization and attribution, LoCD matches each critical span with its corresponding privileged context.
This allows the supervision induced by each privileged context to be applied to the student states where that context is semantically relevant.

\section{Experiments}
\label{sec:experiments}

We evaluate LoCD from three main perspectives.
First, we compare it with strong baselines across reasoning, coding, and agent tasks.
Second, we examine its training dynamics, including downstream performance, policy entropy, and parameter updates throughout training.
Third, we study key design choices and robustness to different distillation objectives.

\subsection{Experimental Setup}

\paragraph{Tasks and models.}
We evaluate LoCD across four settings: math reasoning, code generation,
interactive decision-making, and software-engineering agents.
For math reasoning, we train on a 28.8K subset of
OpenThoughts~\citep{guha2026openthoughts};
for code generation, we use Codeforces problems following the competitive-programming
setting of CodeContests~\citep{li2022competition};
for interactive decision-making, we consider
WebShop~\citep{yao2022webshop} and
ALFWorld~\citep{shridhar2020alfworld};
and for software engineering, we train on SWE-Gym
instances~\citep{pan2024training} and evaluate on a held-out subset of SWE-Gym and SWE-bench Verified~\citep{jimenez2024swe} using mini-swe-agent~\citep{yang2024swe}.

Our main experiments use Qwen3-8B~\citep{yang2025qwen3technicalreport}
in non-thinking mode.
To examine generalization across model families, we
additionally evaluate Falcon3-10B-Instruct~\citep{Falcon3}
on math reasoning.
For software-engineering tasks, we use
Qwen3-Coder-30B-A3B-Instruct~\citep{yang2025qwen3technicalreport}.

\begin{table*}[t]
\centering

\setlength{\tabcolsep}{5.0pt}
\renewcommand{\arraystretch}{1.12}

\resizebox{\textwidth}{!}{
\begin{tabular}{lccccccccc}
\toprule
\textbf{Method}
& \textbf{AIME24}
& \textbf{AIME25}
& \textbf{HMMT25-Feb}
& \makecell{\textbf{HMMT25-Nov}}
& \textbf{Olympiad}
& \textbf{MATH-500}
& \textbf{Minerva}
& \textbf{AMO}
& \textbf{Avg.} \\
\midrule

\multicolumn{10}{l}{\textit{Qwen3-8B}} \\

Base
& 28.06
& 20.83
& 11.11
& 11.67
& 55.34
& 85.80
& 39.34
& 2.33
& 31.81 \\

Text2Grad
& $46.30\std{3.7}$
& $33.52\std{2.4}$
& $21.11\std{1.2}$
& $28.52\std{1.9}$
& $60.24\std{2.6}$
& \second{$91.07\std{1.1}$}
& $38.73\std{0.9}$
& $4.28\std{0.6}$
& $40.47$ \\

TRD
& $47.96\std{4.1}$
& $37.50\std{1.2}$
& $21.39\std{1.3}$
& $26.76\std{1.0}$
& $60.68\std{3.1}$
& $89.33\std{0.7}$
& $40.32\std{0.8}$
& $4.72\std{0.4}$
& $41.08$ \\

OPCD
& $54.26\std{2.8}$
& $38.15\std{1.8}$
& $21.39\std{1.6}$
& \second{$29.17\std{4.3}$}
& $60.93\std{1.7}$
& $89.33\std{1.8}$
& \second{$40.81\std{1.6}$}
& \second{$5.06\std{0.6}$}
& \second{$42.39$} \\

OP$^{2}$SD
& \second{$56.85\std{2.7}$}
& \second{$38.98\std{2.4}$}
& $20.28\std{1.4}$
& $26.94\std{2.0}$
& $60.44\std{2.6}$
& $90.07\std{1.1}$
& $40.07\std{2.2}$
& $4.72\std{0.7}$
& $42.29$ \\

ROSD
& $48.70\std{4.4}$
& $36.57\std{2.0}$
& \second{$23.24\std{1.6}$}
& $24.63\std{1.4}$
& \second{$61.97\std{2.2}$}
& $87.87\std{0.8}$
& $39.09\std{1.2}$
& $4.56\std{0.4}$
& $40.83$ \\

\textbf{LoCD}
& \best{$61.30\std{3.6}$}
& \best{$47.87\std{4.6}$}
& \best{$25.28\std{2.5}$}
& \best{$38.43\std{3.0}$}
& \best{$66.32\std{3.0}$}
& \best{$93.80\std{1.5}$}
& \best{$44.61\std{2.6}$}
& \best{$6.67\std{1.0}$}
& \best{$48.04$} \\

\midrule

\multicolumn{10}{l}{\textit{Falcon3-10B-Instruct}} \\

Base
& $4.44$
& $3.06$
& $0.56$
& $4.44$
& $22.55$
& $51.40$
& $32.35$
& $2.33$
& $15.14$ \\

Text2Grad
& $9.07\std{1.4}$
& $3.33\std{0.7}$
& $1.39\std{0.6}$
& $5.00\std{0.7}$
& $35.31\std{1.7}$
& $64.80\std{0.8}$
& $34.56\std{2.4}$
& $3.00\std{0.5}$
& $19.56$ \\

TRD
& \second{$11.11\std{1.6}$}
& $5.56\std{0.8}$
& \second{$2.50\std{0.7}$}
& $4.44\std{0.6}$
& \second{$37.69\std{1.8}$}
& \second{$72.13\std{0.9}$}
& $38.60\std{2.5}$
& \second{$3.33\std{0.6}$}
& \second{$21.92$} \\

OPCD
& $10.83\std{1.4}$
& \second{$6.39\std{0.7}$}
& $1.67\std{0.6}$
& \second{$5.28\std{0.7}$}
& $35.31\std{1.7}$
& $69.67\std{0.9}$
& $38.60\std{2.3}$
& $2.83\std{0.6}$
& $21.32$ \\

OP$^{2}$SD
& $10.56\std{1.5}$
& $5.28\std{0.7}$
& $2.22\std{0.6}$
& $5.00\std{0.7}$
& $36.35\std{1.7}$
& $70.53\std{0.9}$
& $37.75\std{2.3}$
& $3.17\std{0.5}$
& $21.36$ \\

ROSD
& $10.74\std{1.3}$
& $6.30\std{0.6}$
& $1.39\std{0.5}$
& $4.07\std{0.6}$
& $33.23\std{1.7}$
& $68.73\std{0.8}$
& \second{$39.34\std{2.1}$}
& $3.17\std{0.5}$
& $20.87$ \\

\textbf{LoCD}
& \best{$12.41\std{1.7}$}
& \best{$7.22\std{0.9}$}
& \best{$3.15\std{0.7}$}
& \best{$5.56\std{0.8}$}
& \best{$39.91\std{1.8}$}
& \best{$73.47\std{1.0}$}
& \best{$40.44\std{2.6}$}
& \best{$3.83\std{0.6}$}
& \best{$23.25$} \\

\bottomrule
\end{tabular}
}
\vspace{-1mm}
\caption{Performance on math reasoning benchmarks. Results for train-based methods are reported as mean $\pm$ standard deviation over three independent runs.}
\label{tab:main_results}
\vspace{0mm}

\end{table*}

{
\setlength{\dbltextfloatsep}{8pt plus 2pt minus 2pt}
\begin{table*}[t]
\centering
\setlength{\tabcolsep}{5.0pt}
\renewcommand{\arraystretch}{0.98}

\sbox{\codetabbox}{\resizebox{0.65\textwidth}{!}{
\begin{tabular}{lccccc}
\toprule
\textbf{Method}
& \textbf{CF$_{800\text{-}1200}$}
& \textbf{CF$_{1200\text{-}1600}$}
& \textbf{CF$_{1600+}$}
& \textbf{LCB}
& \textbf{Avg.} \\
\midrule


Base
& $26.47$
& $8.20$
& $0.55$
& $16.71$
& $12.98$ \\

Text2Grad
& $24.02\std{1.2}$
& $8.03\std{1.0}$
& $0.37\std{0.1}$
& $16.72\std{0.8}$
& $12.28$ \\

TRD
& \second{$28.66\std{1.6}$}
& \second{$10.69\std{1.3}$}
& $0.65\std{0.1}$
& \second{$18.19\std{0.8}$}
& \second{$14.55$} \\

ROSD
& $28.09\std{1.5}$
& $9.68\std{2.1}$
& $0.62\std{0.2}$
& $17.10\std{1.1}$
& $13.87$ \\

OPCD
& $27.20\std{1.8}$
& $9.88\std{0.5}$
& \second{$0.70\std{0.2}$}
& $17.06\std{0.2}$
& $13.71$ \\

OP$^{2}$SD
& $25.36\std{1.4}$
& $8.92\std{0.6}$
& \second{$0.70\std{0.2}$}
& $16.84\std{0.2}$
& $12.96$ \\

\textbf{LoCD}
& \best{$30.94\std{1.1}$}
& \best{$12.91\std{1.5}$}
& \best{$0.79\std{0.2}$}
& \best{$24.14\std{2.1}$}
& \best{$17.20$} \\

\bottomrule
\end{tabular}
}}

\begin{minipage}[t]{0.65\textwidth}
\centering
\vspace{0pt}
\usebox{\codetabbox}
\end{minipage}
\hfill
\begin{minipage}[t]{0.34\textwidth}
\centering
\vspace{0pt}
\includegraphics[
    totalheight=\dimexpr\ht\codetabbox+\dp\codetabbox\relax
]{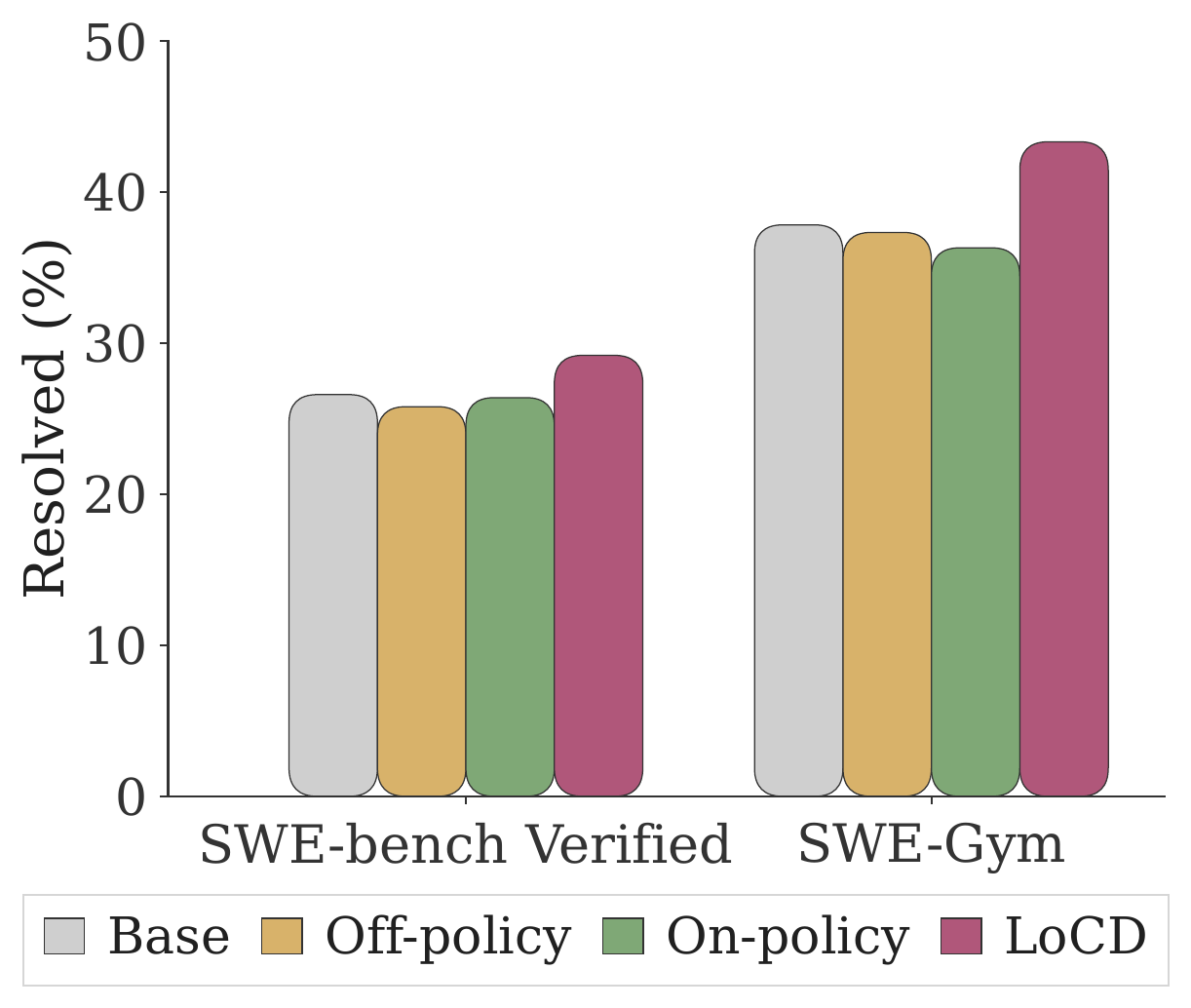}
\vspace{-1mm}
\end{minipage}

\par\vspace{0.25em}

\begin{minipage}[t]{0.65\textwidth}
\centering
\captionof{table}{Performance on code-generation benchmarks.}
\label{tab:main_results_code}
\end{minipage}
\hfill
\begin{minipage}[t]{0.33\textwidth}
\centering
\captionof{figure}{Performance on SWE.}
\label{fig:swe_bars}
\vspace{-1mm}
\end{minipage}
\vspace{-5mm}
\end{table*}}

\paragraph{Baselines.}
We compare LoCD with representative approaches for learning from interaction experience.
Text2Grad~\citep{wang2026text2grad} converts free-form language feedback into
span-level learning signals for policy optimization.
TRD~\citep{jiang2026trajectory} uses teacher guidance to revise student-generated
trajectories before distillation, providing a strong off-policy alternative.
For on-policy context distillation,
OPCD~\citep{ye2026policy} distills a context-conditioned teacher on
student-generated trajectories;
ROSD~\citep{zhao2026rosd} identifies the first erroneous region and restricts
distillation to the affected suffix;
and OP$^2$SD~\citep{ichihara2026privileged} uses shuffled privileged solutions to induce alternative teacher behavior while retaining student-generated rollouts (requiring reference solutions unavailable in our agent tasks).
For multi-step agent tasks, we additionally compare with Hint-SD~\citep{yeo2026hint}, which applies
targeted hindsight supervision to failure-relevant parts of an agent trajectory.
All methods use the same training datasets, feedback sources,
and experience-construction protocols, while collecting
trajectories from their respective evolving policies.
For each task, we use the same number of training steps
and batch size across methods.

\paragraph{Training and evaluation.}
For math reasoning, we use reference solutions as
feedback; for code generation, we use official editorials
and execution feedback.
For agent tasks, feedback is derived from environment
observations, execution results, and task outcomes,
without relying on expert demonstrations.
LoCD iteratively collects new trajectories from the updated
student and performs feedback-guided localized distillation.
We use reverse KL as the default distillation objective
for LoCD and on-policy context-distillation baselines.

For evaluation, we report avg@12 on AIME, HMMT, and AMO,
and avg@1 on MATH-500, OlympiadBench, and Minerva.
For code generation, we report average success rates over
12 samples per problem on Codeforces and LiveCodeBench.
For interactive agents, we report task success rates using
avg@1 on WebShop and avg@3 on ALFWorld.
For software engineering, we report resolved rates using
avg@4 on SWE-Gym and pass@1 on SWE-bench Verified.
Additional details on dataset construction, feedback
generation, training configurations, and evaluation
protocols are provided in Appendix~\ref{app:experimental_details}.

\subsection{Main Results}
\vspace{-1mm}
Tables~\ref{tab:main_results} and~\ref{tab:main_results_code}
show that LoCD consistently outperforms strong baselines
on both math reasoning and code generation.
On Qwen3-8B, LoCD achieves the best results across all
eight math benchmarks, improving the average score from
42.39 to 48.04, with consistent improvements also observed
on Falcon3-10B-Instruct.
For code generation, LoCD achieves an average solve rate
of 17.20\%, compared with 14.55\% for the strongest baseline.
Notably, the gains are particularly pronounced on challenging
benchmarks such as AIME25, HMMT25-Nov, and LiveCodeBench,
consistent with our motivation to introduce corrective
behaviors that are difficult for the student to generate.

Beyond single-step reasoning and coding tasks, LoCD also
demonstrates strong performance on multi-step agent tasks.
As shown in Table~\ref{tab:agent_main}, LoCD outperforms
the strongest baselines by 5.60 and 7.46 percentage points
on WebShop and ALFWorld, respectively.
Figure~\ref{fig:swe_bars} shows similar improvements on
SWE-bench Verified and SWE-Gym, where LoCD achieves the
highest resolved rates among the compared methods.
These results further demonstrate the effectiveness of LoCD
across different forms of interaction feedback, including
environment observations and execution failures, without
requiring reference solution.
We further examine whether these improvements persist under
larger sampling budgets, detailed in Appendix~\ref{app:passk}.

\begin{table*}[t]
\centering

\setlength{\tabcolsep}{5.0pt}
\renewcommand{\arraystretch}{1.06}

\resizebox{0.92\textwidth}{!}{
\begin{tabular}{lccccccccccc}
\toprule
& \multicolumn{4}{c}{\textbf{WebShop (official test)}}
& \multicolumn{7}{c}{\textbf{ALFWorld (valid\_unseen)}} \\
\cmidrule(lr){2-5}
\cmidrule(lr){6-12}

\textbf{Method}
& $C_1$
& $C_2$
& $C_{3+}$
& \textbf{All}
& \textbf{Pick}
& \textbf{Look}
& \textbf{Clean}
& \textbf{Heat}
& \textbf{Cool}
& \textbf{Pick2}
& \textbf{All} \\
\midrule

Base
& $58.24$
& $8.96$
& $9.22$
& $18.00$
& $27.78$
& \second{$27.78$}
& $18.28$
& \second{$31.88$}
& $26.98$
& $1.96$
& $22.89$ \\

TRD
& $63.74$
& $43.28$
& \second{$30.50$}
& $43.40$
& $15.28$
& $20.37$
& $19.35$
& $21.74$
& $22.22$
& $9.80$
& $18.41$ \\

OPCD
& $62.64$
& $15.67$
& $12.77$
& $23.40$
& \second{$29.17$}
& $24.07$
& $20.43$
& $28.99$
& $28.57$
& \second{$13.73$}
& $24.38$ \\

ROSD
& $62.64$
& $47.39$
& $26.95$
& $44.40$
& $26.39$
& $25.93$
& \second{$23.66$}
& \second{$31.88$}
& \second{$30.16$}
& \second{$13.73$}
& \second{$25.62$} \\

Hint-SD
& \second{$65.93$}
& \second{$48.88$}
& \second{$30.50$}
& \second{$46.80$}
& $25.00$
& $24.07$
& \second{$23.66$}
& $18.84$
& $28.57$
& \best{$15.69$}
& $22.89$ \\

\textbf{LoCD}
& \best{$69.23$}
& \best{$56.72$}
& \best{$33.33$}
& \best{$52.40$}
& \best{$33.33$}
& \best{$35.19$}
& \best{$26.88$}
& \best{$49.28$}
& \best{$36.51$}
& \best{$15.69$}
& \best{$33.08$} \\

\bottomrule
\end{tabular}
}
\vspace{-1mm}
\caption{Performance on interactive decision-making benchmarks.}
\label{tab:agent_main}
\vspace{-1mm}
\end{table*}

\begin{figure}[t]
    \centering
    \includegraphics[width=\linewidth]{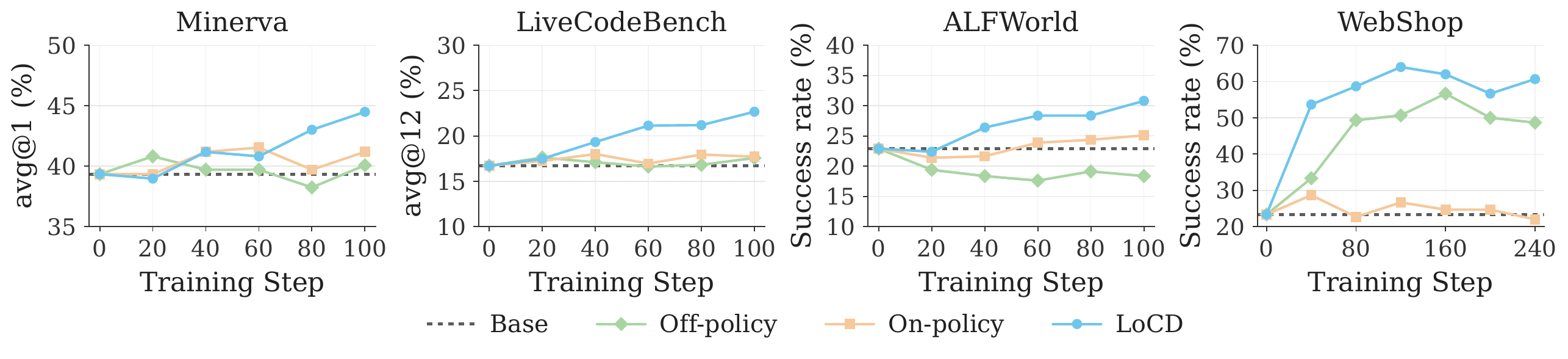}
    \vspace{-6mm}
    \caption{Training dynamics across four representative tasks.}
    \label{Training dynamics across four representative tasks.}
    \vspace{-1mm}
\end{figure}

\begin{figure}[t]
\centering
\includegraphics[width=\linewidth]{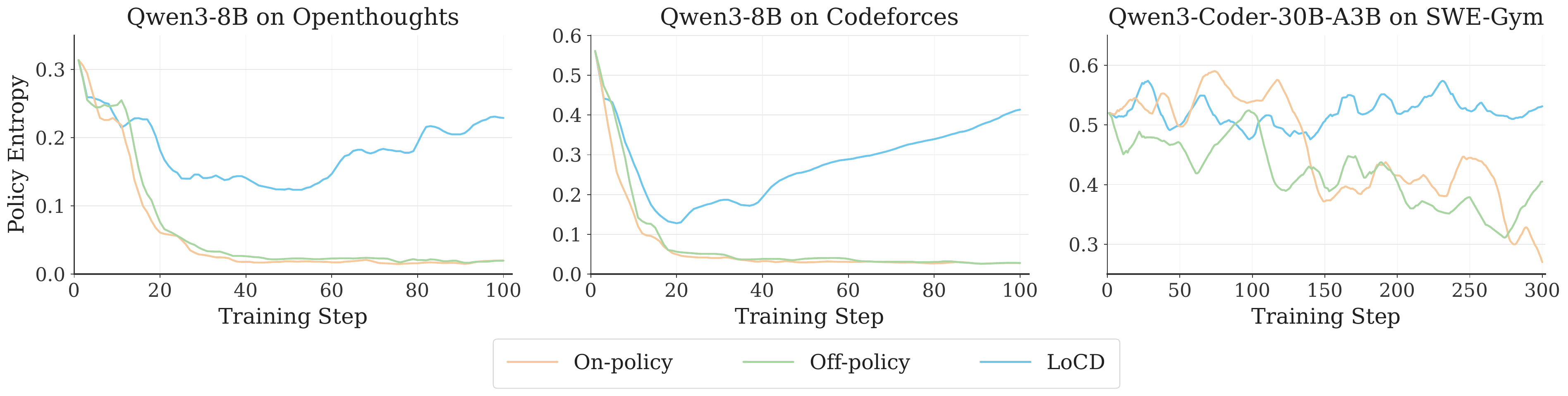}
\vspace{-5mm}
\caption{Policy entropy during training on OpenThoughts, Codeforces, and SWE-Gym.}
\label{fig:entropy}
\vspace{-1mm}
\end{figure}

\vspace{-2mm}
\subsection{Training Dynamics}
\label{sec:training_dynamics}
\vspace{-1mm}
We then study how task performance, policy entropy, and model parameters evolve during training.
In the following analyses, we use TRD and OPCD as the off-policy and on-policy context-distillation baselines, respectively.
\vspace{-2mm}
\paragraph{Performance over training.}
Figure~\ref{Training dynamics across four representative tasks.} compares LoCD with on-policy and off-policy
context distillation throughout training.
Across all four representative tasks, LoCD continues to improve as training
progresses, whereas the two baselines tend to plateau or fluctuate.
The separation is particularly pronounced on WebShop and ALFWorld, where LoCD continues to gain performance after the baselines have largely saturated.

\paragraph{Policy entropy.}
We measure policy entropy as the average token-level
entropy over training rollouts.
Figure~\ref{fig:entropy} records how the output distribution evolves during training.
Across OpenThoughts, Codeforces, and SWE-Gym, both on-policy and off-policy
context distillation progressively reduce policy entropy.
In contrast, LoCD avoids this sustained entropy collapse and maintains a more
stable entropy level throughout training.
Together with its stronger downstream performance, this indicates that LoCD can
improve task behavior without excessively concentrating the model's output
distribution.

\begin{figure}[t]
\centering
\includegraphics[width=\linewidth]{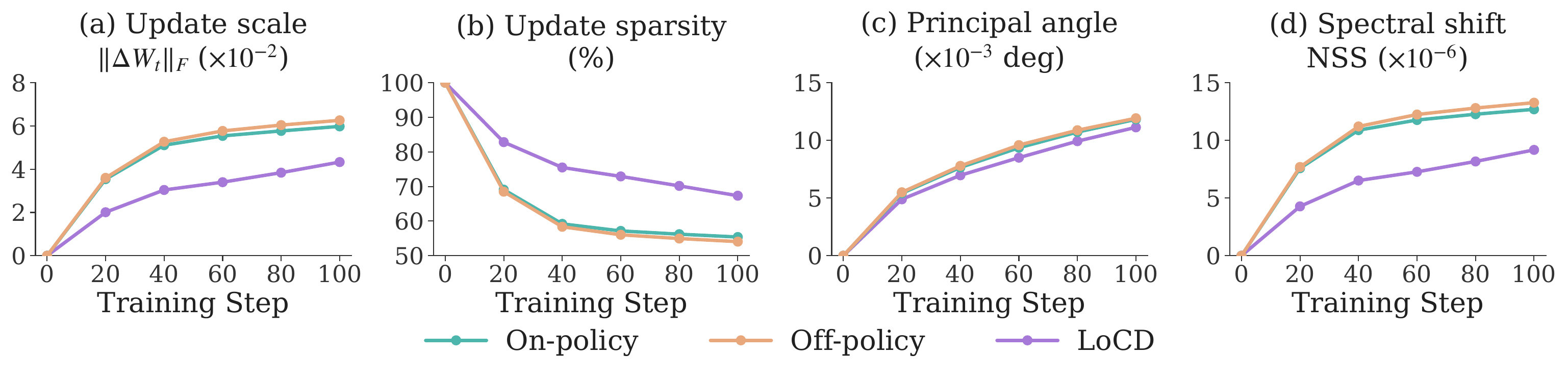}
\vspace{-6mm}
\caption{Parameter-update geometry during math reasoning training.}
\vspace{-1mm}
\label{Parameter-update geometry during math reasoning training.}
\end{figure}

\paragraph{Parameter-update geometry.}
We further examine how different distillation objectives
modify model parameters during training.
As shown in Figure~\ref{Parameter-update geometry during math reasoning training.},
LoCD exhibits smaller cumulative parameter changes
and higher visible update sparsity than on-policy and
off-policy context distillation.
Similar patterns are observed on code generation and
ALFWorld, with detailed analyses provided in Appendix~\ref{app:geometry}.

\begin{figure}[t]
\centering
\begin{minipage}[t]{0.40\textwidth}
    \centering
    \includegraphics[width=\linewidth]{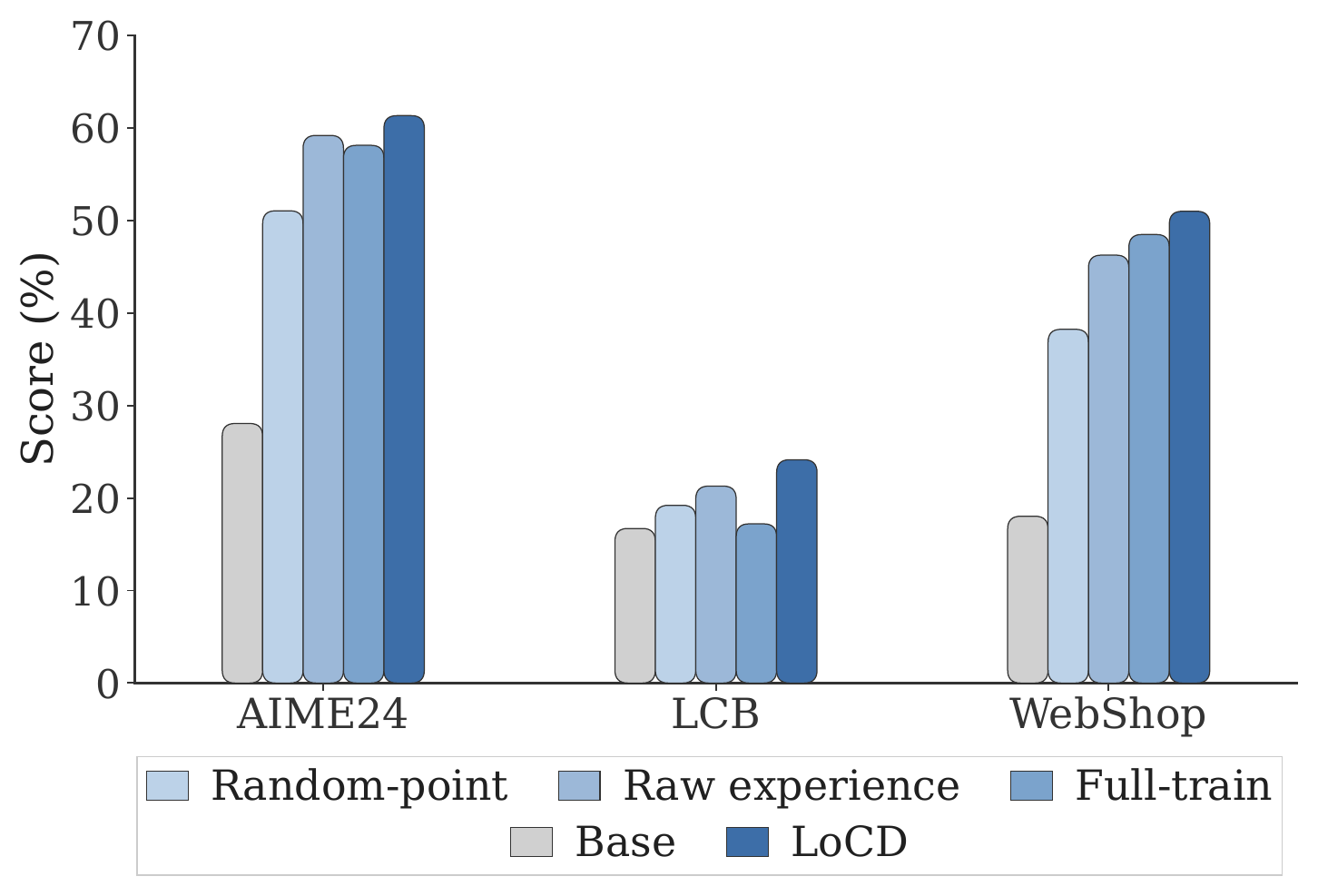}
    \captionof{figure}{Ablation of LoCD.}
    \label{fig:Ablation of LoCD.}
\end{minipage}
\hfill
\begin{minipage}[t]{0.56\textwidth}
    \centering
    \includegraphics[width=\linewidth]{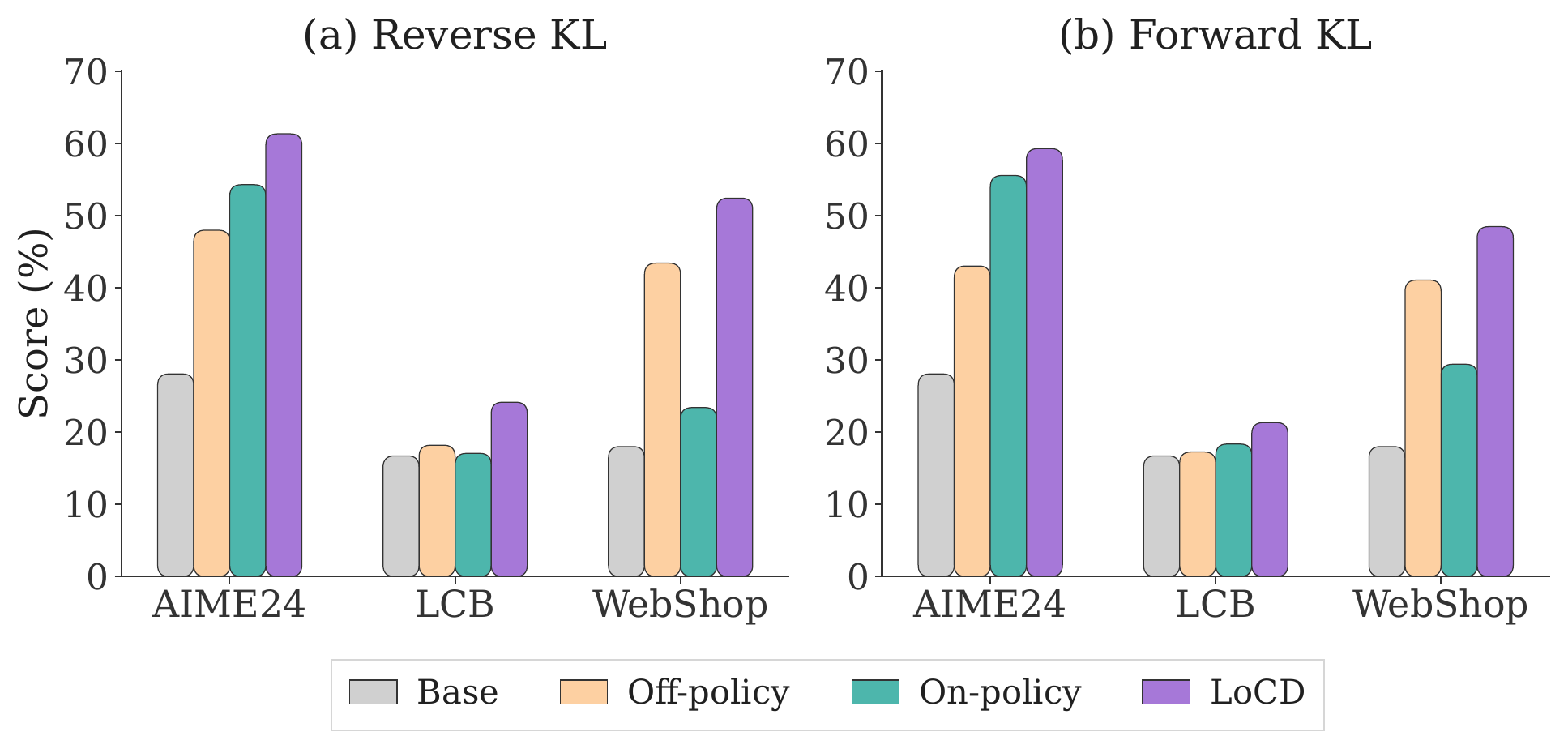}
    \captionof{figure}{Robustness to divergence choice.}
    \label{fig:kl_robustness}
\end{minipage}
\vspace{-1mm}
\end{figure}

\subsection{Ablation Study}

\paragraph{Ablation of LoCD.}
We examine three key design choices in LoCD.
\emph{Random Point} replaces feedback-guided localization
with a randomly selected correction point.
\emph{Raw Experience} uses the complete interaction experience
as privileged context instead of extracting a targeted hint.
\emph{Full-Train} retains the original student prefix
but extends teacher generation and distillation to the
entire remaining trajectory rather than a localized span.

As shown in Figure~\ref{fig:Ablation of LoCD.}, all three
variants consistently underperform LoCD across the evaluated
benchmarks.
The degradation from \emph{Random Point} and \emph{Raw Experience} suggests
that effective supervision requires both identifying
failure-relevant states and extracting targeted corrective
information.
Notably, \emph{Full-Train} also performs worse despite providing
supervision over a longer teacher-generated continuation,
highlighting the importance of restricting distillation
to critical spans.

\paragraph{Robustness to divergence choice.}
We further examine whether the effectiveness of LoCD depends on the choice of distillation divergence.
Figure~\ref{fig:kl_robustness} compares forward and reverse KL on AIME24, LiveCodeBench~v6, and WebShop.
Although the relative performance of the on-policy and off-policy baselines varies with the divergence direction, LoCD consistently achieves the strongest performance under both choices.
Forward and reverse KL weight next-token discrepancies by teacher and student probabilities, respectively. LoCD provides informative teacher-generated correction states anchored to student prefixes under either objective, which may explain its consistent gains across both weighting schemes.

\section{Related Works}
\paragraph{Non-Parametric Learning from Experience.}
A broad line of work reuses past interaction experience without updating model
parameters. Some methods directly store and retrieve past trajectories,
feedback, or reflections as in-context examples
\citep{shinn2023reflexion,zhao2024expel,liu2025contextual}.
More recent methods abstract accumulated interactions into reusable insights,
guidelines, skills, or workflows that generalize across related tasks
\citep{wang2023voyager,fu2024autoguide,wang2024agent,yang2025learning}.
These methods rely on retrieving and reintroducing relevant experience at
inference time, whereas we study how to internalize such experience directly
into model parameters
\citep{liu2025contextual,wang2024agent,cao2026remember}.

\paragraph{Reinforcement Learning from Interaction Experience.}
Reinforcement learning has become a central paradigm for improving large language models, typically optimizing scalar rewards derived from preference models or outcome verifiers
\citep{ouyang2022training,shao2024deepseekmath}.
However, interaction feedback often contains richer diagnostic and corrective information expressed in natural language than scalar rewards can capture
\citep{wang2026text2grad,zhou2025sequence,wang2026reward}.
Recent work therefore incorporates language feedback into reinforcement learning by converting critiques into token-, span-, or process-level rewards, or by using feedback to guide refinement and exploration
\citep{cao2024enhancing,wang2026text2grad,zhang2025critique,li2025lanpo}.
Related approaches improve credit assignment by localizing optimization to critical steps or error regions, or introducing localized interventions during policy optimization
\citep{li2026verified,liang2026learning,li2026ssvpo,huang2026learning,chen2026mintrl}.
Despite these advances, natural-language feedback is still largely mediated through reward or credit-assignment signals, requiring the model to discover the corresponding corrective behavior through policy optimization.

\paragraph{Context Distillation.}
Context distillation internalizes behaviors induced by additional context into
model parameters by training a student without that context to match a
context-conditioned teacher
\citep{askell2021general,snell2022learning}.
In learning from interaction feedback, off-policy approaches typically use
privileged information to construct improved responses and then train the
student on these corrections
\citep{scheurer2023training}.
InT~\citep{yang2026int} constructs targeted corrections offline and uses SFT to improve reasoning before subsequent reinforcement learning, whereas LoCD dynamically generates corrections from the current student's prefixes
throughout training.
Recent on-policy methods instead sample trajectories from the current student and provide privileged teacher supervision on student-generated prefixes,
allowing reference solutions, demonstrations, or textual feedback to be internalized while remaining aligned with the student's state distribution
\citep{zhao2026self,shenfeld2026self,hubotter2026reinforcement,ye2026policy}.
More recent work extends this paradigm toward self-evolving models by deriving privileged context from interaction experience, including transferable insights, hindsight skills, and learnable latent contexts
\citep{ren2026self,wu2026seed,zhang2026latent}.
Other work further improves the alignment between privileged feedback and student states by constructing locally relevant hindsight feedback or targeting failure-relevant actions for distillation~\citep{liu2026hero,yeo2026hint,feng2026past}.
Relay-OPD~\citep{xu2026pass} triggers temporary teacher takeovers based on disagreements over predefined reflection tokens and distills the resulting mixed trajectories using a separate stronger teacher.
Despite these advances, effectively introducing corrective behaviors beyond the student's own rollouts while limiting off-policy distribution shift remains challenging.
LoCD addresses this challenge by using interaction feedback to localize critical spans, regenerating them from student prefixes with a context-conditioned version of the same model, and restricting distillation to the corrected spans.

\section{Conclusion}
We study how language models can internalize interaction experience through context distillation.
We find that on-policy context distillation becomes inefficient when the required behavior is difficult for the student to produce.
Motivated by this limitation, LoCD introduces localized teacher corrections from student-generated prefixes, enabling effective learning of new behaviors while limiting off-policy distribution shift.
Extensive experiments across math reasoning, code generation, and agent tasks demonstrate that LoCD shows  consistent improvements over strong baselines and maintains more stable policy entropy with smaller and sparser parameter updates.





\bibliography{iclr2027_conference}
\bibliographystyle{iclr2027_conference}
\clearpage
\appendix

\startcontents[appendix]
\section*{Appendix Contents}
\printcontents[appendix]{}{1}{}

\clearpage

\section{Details of Empirical Study}
\label{Details of Empirical Study}

\subsection{Details of the Initial-Accuracy Study}
\label{Details of the Initial-Accuracy Study}

For each model, we construct a single-step interaction experience for each
\textsc{OpenThoughts} problem.
Specifically, we sample a student output $a_1$, use the reference solution as
feedback $f_1$, and form the interaction trajectory
$\tau=(a_1,f_1)$.
We then use the same model as the experience extractor $\mathcal{E}$ to obtain
the privileged context
$c\sim\mathcal{E}(\cdot\mid x,\tau)$.
For each problem, we estimate the student's initial accuracy using 128
independent rollouts and assign it to one of six bins:
$\{0\}$, $(0,20]$, $(20,40]$, $(40,60]$, $(60,80]$, and $(80,100]$.
The three models are grouped independently according to their own initial
accuracy, and we sample 100 problems from each bin for training.

For each model and accuracy bin, we train OPCD and TRD for 5 epochs as the
on-policy and off-policy context-distillation methods, respectively.
Both methods are trained on the same sampled problems and their corresponding
privileged context.
After training, we evaluate the distilled student without privileged context
on the same problems using avg@128.
All other training and evaluation settings follow those used in our main
experiments.

\subsection{Details of the Controlled Knowledge-Injection Study}
\label{app:knowledge_injection}

We construct a synthetic knowledge-injection task to isolate the learning
dynamics of context distillation from other factors.
Each example follows a single factual template,
``\textit{$X$'s $R$ is $Y$},'' with the corresponding query
``\textit{What is $X$'s $R$?}''
Both the subject $X$ and the words composing the answer $Y$ are randomly
generated pseudo-words and $R$ is a random relationship.
We generate these pseudo-words using a syllable-based grammar and filter out
real English words to minimize overlap with the model's prior knowledge.
We use a single relation and query template throughout the dataset to avoid
variation in relation difficulty.
In addition, every pseudo-word appears only once across the entire dataset, so
different facts share no subject or answer words.
The synthetic fact is directly provided to the privileged teacher as context,
while the student receives only the query.

To control the length of the behavior to be learned, we construct answers whose
tokenized lengths are exactly
$L\in\{4,8,16,32\}$.
Specifically, each answer is formed by concatenating pseudo-words and retaining
only combinations whose tokenization contains exactly $L$ tokens at the
generation position.
No padding or rewriting is used.
All length groups are generated from the same pseudo-word distribution and are
mixed within the same training run, making answer length the primary controlled
variable.
We conduct this study with Qwen3-4B using the same on-policy and off-policy
context distillation procedures as above, training for 200 epochs and evaluating
every 5 epochs.
Each length group contains eight facts.
At evaluation time, the student receives no privileged context and independently
generates eight responses for each fact.
We report exact-match accuracy averaged first over the eight generations for
each fact and then over all facts in the corresponding length group.
All other training and evaluation settings follow the main experiments.

\section{What Does On-Policy Context Distillation Really Learn?}
\label{app:pi_semantics}

On-policy context distillation improves student performance
by distilling behaviors induced by privileged context.
However, does such improvement necessarily indicate
that the student has internalized the task-specific knowledge
provided by that context?
To investigate this question, we conduct two complementary
experiments examining the role of privileged context
in on-policy context distillation.

\subsection{Performance Gains with Task-Irrelevant Privileged Context}
\label{app:pi_semantic_ablation}

\paragraph{Experimental setup.}
We investigate whether the benefits of on-policy context
distillation depend on the task-specific correctness of
privileged context.
Using Qwen3-8B, we replace the
correct reference solution with three alternative forms
of privileged context:
(1) \emph{Shuffled Solutions}, randomly sampled solutions
to other problems;
(2) \emph{Few-shot Exemplars}, high-quality reasoning
demonstrations;
and (3) \emph{Wrong Solutions}, solutions that do not
correctly solve the current problem.
We evaluate the resulting models on AIME24, AIME25,
and HMMT25.

\begin{figure}[t]
    \centering
    \includegraphics[width=\linewidth]{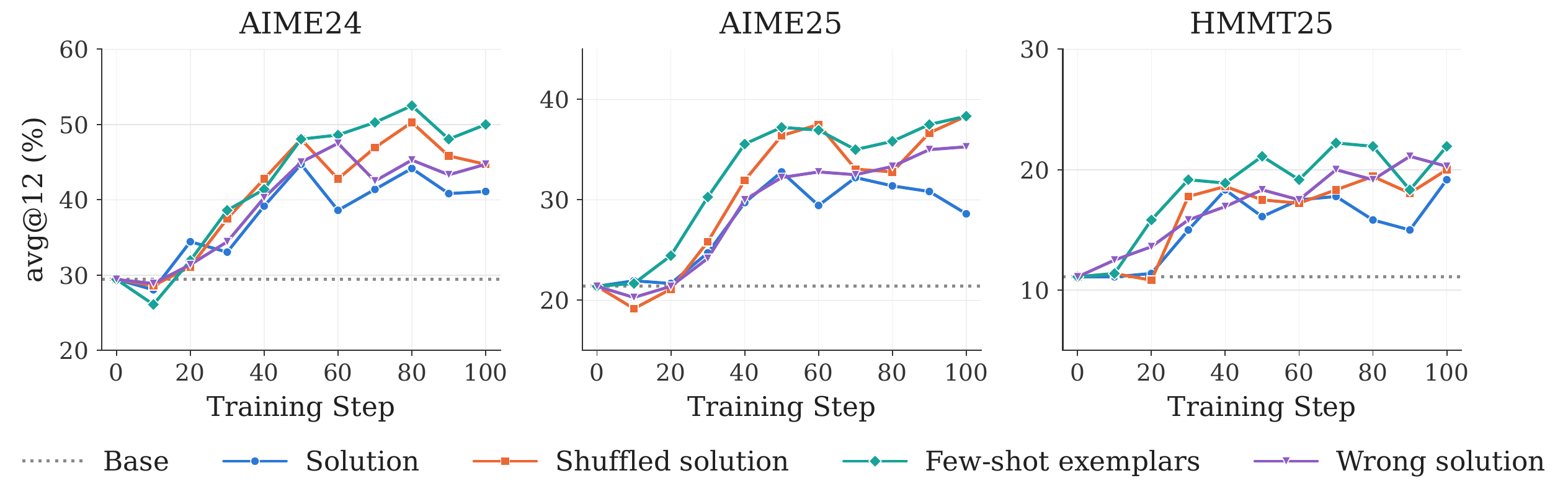}
    \caption{
        On-policy context distillation with different
        forms of privileged context on Qwen3.
        Substantial performance improvements persist
        even when the privileged context does not provide
        the correct solution to the current problem.
    }
    \label{fig:pi_semantic_ablation}
\end{figure}

\paragraph{Results.}
As shown in Figure~\ref{fig:pi_semantic_ablation},
on-policy context distillation still yields substantial
performance improvements even when privileged context
does not contain the correct solution to the current problem.
In some settings, shuffled solutions or few-shot exemplars
even outperform the correct reference solutions.
These results suggest that performance gains from
on-policy context distillation do not necessarily
reflect the internalization of task-specific knowledge.

\subsection{Generic Instructions and Entropy Reduction}
\label{app:pi_behavioral_instructions}

\paragraph{Experimental setup.}
We further examine whether on-policy context distillation
can improve performance without additional task-specific
knowledge.
We replace reference solutions with three generic
behavioral instructions: \emph{Brief}, \emph{Decisive},
and \emph{Critical}.
These instructions contain no task-specific answers or
additional factual knowledge.

\paragraph{Prompt templates.}
All three settings share the same prompt template,
differing only in the instruction provided.

\begin{lstlisting}[
    basicstyle=\ttfamily\small,
    breaklines=true,
    columns=fullflexible,
    frame=single
]
Problem: {problem}

=== Instruction Begin ===
{instruction}
=== Instruction End ===
Now solve the problem above, following the instruction. Reason step by step, and put your final answer within \boxed{}.
\end{lstlisting}

The instructions are defined as follows:

\begin{description}
    \item[\emph{Brief:}]
    Please reduce unnecessary reasoning and give your
    final answer as quickly as possible.

    \item[\emph{Decisive:}]
    Commit to a single approach and carry it through
    to the end. Do not backtrack or restart.

    \item[\emph{Critical:}]
    Question every step. Consider alternative approaches
    and revisit earlier steps before continuing.
\end{description}

\begin{figure}[t]
    \centering
    \includegraphics[width=\linewidth]{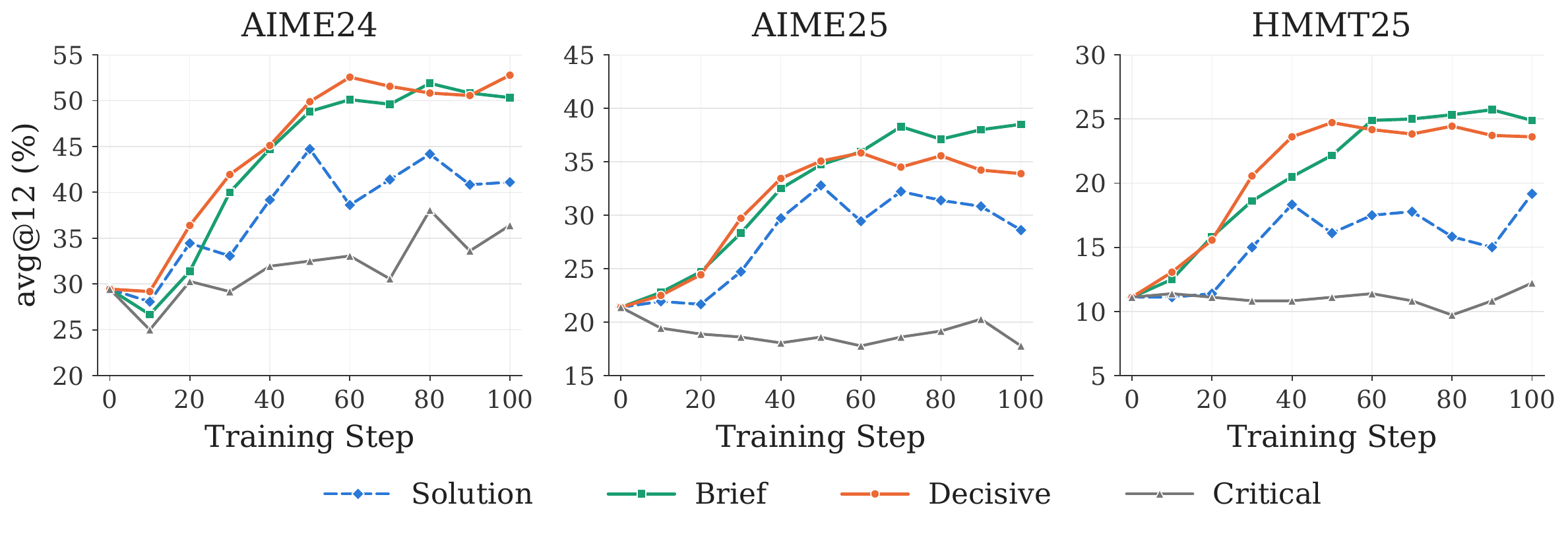}
    \caption{
        Performance of on-policy context distillation
        with different generic behavioral instructions.
        Simple instructions can yield substantial improvements
        without providing task-specific knowledge.
    }
    \label{fig:pi_behavioral_instructions}
\end{figure}

\begin{figure}[t]
    \centering
    \includegraphics[width=\linewidth]{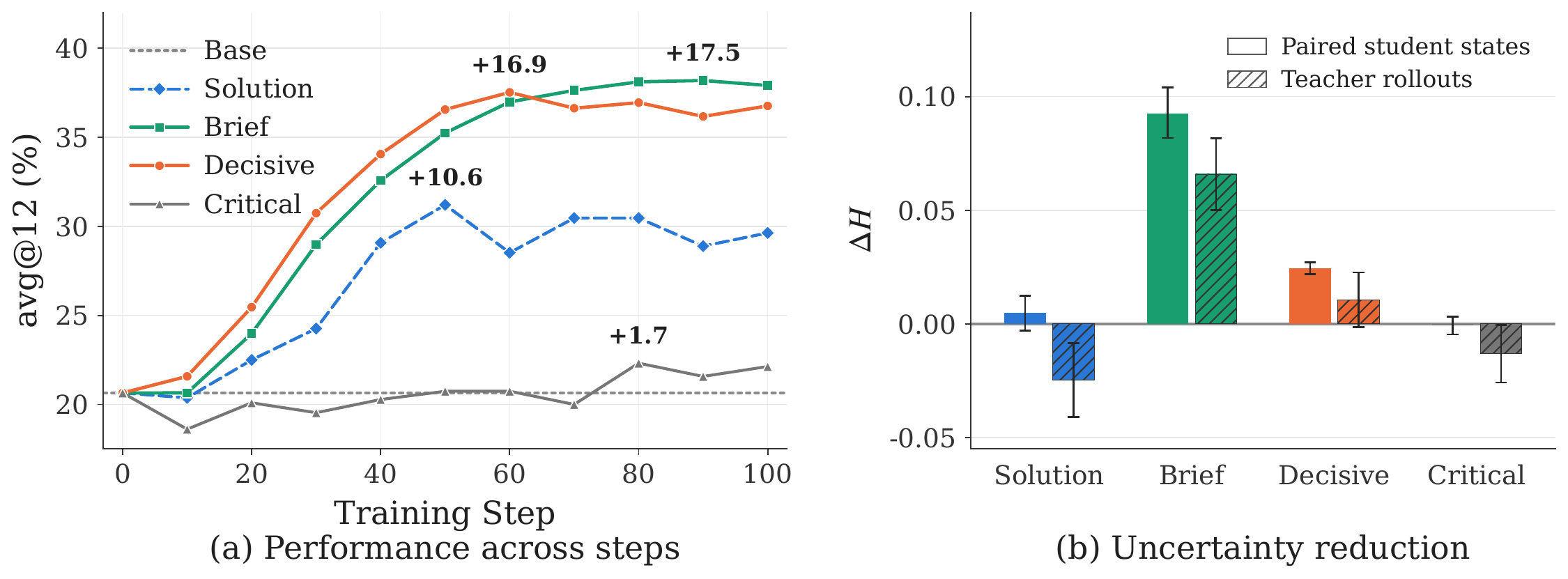}
    \caption{
        Mean performance gains and entropy reduction under
        different privileged contexts.
        (a) Accuracy throughout training.
        (b) Entropy reduction evaluated on student-generated
        prefixes (paired student states) and teacher-generated
        prefixes (teacher rollouts).
        Larger entropy reductions are generally associated
        with greater performance improvements.
    }
    \label{fig:pi_uncertainty}
\end{figure}

\paragraph{Results.}
As shown in Figure~\ref{fig:pi_behavioral_instructions},
these generic instructions yield substantial performance
improvements, sometimes surpassing those obtained with
correct reference solutions.
Figure~\ref{fig:pi_uncertainty}(a) further shows that
\emph{Brief} and \emph{Decisive} improve mean accuracy (AIME24, AIME25, HMMT25) by
17.5 and 16.9 percentage points, respectively,
compared with 10.6 points using correct reference
solutions and only 1.7 points with \emph{Critical}.

To better understand these differences, we examine
how privileged context affects the teacher's output
entropy under two state distributions.
For \emph{paired student states}, we compare the output
entropy with and without privileged context at the same
student-generated prefixes.
For \emph{teacher rollouts}, we perform the comparison
at prefixes generated by the context-conditioned teacher.
We report the entropy reduction $\Delta H$, where
positive values indicate lower entropy when conditioning
on privileged context.

As shown in Figure~\ref{fig:pi_uncertainty}(b),
\emph{Brief} and \emph{Decisive} induce substantial
entropy reductions under both state distributions.
In contrast, correct reference solutions produce
little entropy reduction, while \emph{Critical}
shows little or no reduction.
The correspondence between entropy reduction and
performance improvement suggests that reducing
the teacher's output uncertainty may be an important
source of gains in on-policy context distillation,
beyond the transfer of task-specific knowledge.

\paragraph{Takeaway.}
Together, these findings suggest that performance gains
from on-policy context distillation do not necessarily
reflect the internalization of task-specific privileged
knowledge.
Such gains may also be associated with changes in the
teacher's output distribution, particularly entropy
reduction.
This motivates the central question examined in
Section~\ref{Empirical Study}:
whether on-policy context distillation can effectively
internalize new knowledge that the student does not
already possess.

\section{Experimental Details}
\label{app:experimental_details}

In this section, we provide the data preparation, feedback construction, training configurations, and evaluation protocols for the main experiments in Section~\ref{sec:experiments}.

\subsection{Datasets and Benchmarks}
\label{app:datasets}

\paragraph{Math reasoning.}
We train on a 28.8K-example subset of OpenThoughts~\citep{guha2026openthoughts}. We evaluate on eight benchmarks: AIME24, AIME25, HMMT25-Feb, HMMT25-Nov, OlympiadBench, MATH-500, Minerva, and AMO. The evaluation sizes and sampling protocols are specified in Appendix~\ref{app:evaluation_protocols}.

\paragraph{Code generation.}
We construct the training set from \texttt{open-r1/codeforces}. We first retain problems with difficulty ratings in $[800,1600)$, non-empty official editorials, and standard-input (\texttt{stdio}) interfaces, yielding 2120 candidates. We further retain only problems for which R1-generated reference solutions pass our execution-based judge, resulting in 1793 training problems. We evaluate on CF-422, comprising 422 Codeforces problems disjoint from the training set, and on LiveCodeBench v6. The CF-422 evaluation set includes nine problems without difficulty ratings; the remaining problems are organized into difficulty groups for reporting.

\paragraph{WebShop.}
We use the official WebShop~\citep{yao2022webshop} splits, comprising 9,464 training tasks, 886 development tasks, and 500 test tasks. A subset of 150 development tasks is used for checkpoint selection, and the complete 500-task test set is used for final evaluation.

\paragraph{ALFWorld.}
We construct separate training and development subsets from the official ALFWorld~\citep{shridhar2020alfworld} training directory, containing 1842 and 404 tasks, respectively. We use 150 development tasks for checkpoint selection. The official \texttt{valid\_seen} and \texttt{valid\_unseen} evaluation sets contain 140 and 134 tasks, respectively; the main results are reported on \texttt{valid\_unseen}.

\paragraph{Software engineering.}
We train on 1293 instances selected from the official SWE-Gym~\citep{pan2024training} training set. Our SWE-Gym evaluation includes 265 instances: 148 in-domain instances drawn from eight repositories represented in training, and 117 out-of-distribution instances from three other repositories (\texttt{modin}: 55; \texttt{hydra}: 39; \texttt{bokeh}: 23). The in-domain and out-of-distribution designations refer to repository overlap, not instance overlap: evaluation instances are disjoint from the training instances. We additionally evaluate on all 500 instances of the official SWE-bench Verified benchmark. Both software-engineering evaluations use Mini-SWE-agent as the agent scaffold.

\subsection{Feedback Construction}
\label{app:feedback_construction}

For each task, the current student produces a trajectory and receives task-specific feedback. We use this interaction experience to identify a failure-relevant span or step and produce a corresponding privileged hint. The localization and hint-generation procedure uses the current model with greedy decoding. The resulting hint is provided only in the privileged context for corrective generation and distillation, not to the student at inference time.

\paragraph{Math reasoning.}
For mathematical reasoning, the attribution procedure receives the student's original solution together with the reference solution. The model compares the two to identify an erroneous reasoning span and generate a corresponding corrective hint.

\paragraph{Code generation.}
For Codeforces, we use the official problem editorial as reference feedback. We additionally execute student-generated programs against the problem's tests and checker, obtaining a correctness verdict and, for unsuccessful submissions, a failure category such as wrong answer, runtime error, or timeout. This information is provided for error localization and hint generation.

\paragraph{WebShop.}
For each failed episode, the attribution procedure is invoked once with the customer instruction, the complete numbered observation--action trajectory, and terminal feedback. The instruction specifies the requested product type, attributes, options, and price limit. Terminal feedback includes the reason for termination (e.g., purchase, missing action, context truncation, or step limit) and, if a purchase was made, the simulator's score and an itemized requirement check derived from its built-in reward function. This check indicates whether the purchased item meets the requested type, price constraint, attributes, and selected options, but does not provide the correct product or action. The hint identifies unmet requirements without directly supplying a revised action. No expert demonstrations are used.

\paragraph{ALFWorld.}
For each failed episode, the attribution procedure receives the task description, complete observation--action trajectory, and episode outcome. The trajectory contains simulator observations, including feedback on invalid actions (e.g., ``Nothing happens.''), and the terminal information identifies unsuccessful completion and invalid-action steps. Unlike WebShop, ALFWorld does not provide an itemized reward breakdown. The hint is restricted to objects and receptacles already observed in the trajectory; no expert trajectory or reference action is supplied.

\paragraph{Software engineering.}
For SWE-Gym, the interaction experience includes the agent's tool-call trajectory, tool execution results, test outcomes, and concrete error messages. These execution traces and test feedback are used for localization and corrective hint generation.

\subsection{Training Configurations}
\label{app:training_configs}

All methods use full-parameter fine-tuning with PyTorch AdamW and FSDP. We use a constant learning rate of $1\times10^{-6}$ without warmup or decay, Adam coefficients $(\beta_1,\beta_2)=(0.9,0.999)$, weight decay $0.01$, numerical stability parameter $\epsilon=10^{-8}$, and global gradient-norm clipping at $0.5$. For a given task, LoCD and all baselines use the same batch size and number of optimization steps, as well as the same task data and feedback-construction protocol. The methods collect trajectories from their respective evolving policies, so their sampled trajectories need not be identical. These controls do not imply identical total compute across methods.

\begin{table}[t]
\centering
\caption{Training configurations for the main experiments. The sampling temperature is shared by the initial student rollout and hint-conditioned correction generation.}
\label{tab:appendix_training_configs}
\small
\setlength{\tabcolsep}{4pt}
\renewcommand{\arraystretch}{1.12}
\begin{tabular}{lcccc}
\toprule
\textbf{Task} & \textbf{Steps} & \textbf{Batch} & \textbf{Temp.} & \textbf{Ckpt. interval} \\
\midrule
Math & 100 & 32 & 1.1 & 20 \\
Codeforces & 100 & 32 & 1.1 & 20 \\
WebShop & 240 & 32 & 1.0 & 40 \\
ALFWorld & 100 & 32 & 1.0 & 20 \\
SWE-Gym & 300 & 16 & 1.0 & 50 \\
\bottomrule
\end{tabular}
\end{table}

During training, the student rollout is limited to 4096 new tokens for mathematics and 2048 for Codeforces. WebShop and ALFWorld use a per-step generation limit of 512 tokens. For mathematical reasoning, each batch contains 32 prompts with one student rollout per prompt. After each optimization update, LoCD collects new student trajectories and repeats the attribution, correction-generation, and distillation procedure. Successful rollouts are supervised with SFT over the entire original generated trajectory, while failed rollouts undergo method-specific feedback-based learning; this treatment is shared by LoCD and the baselines.
Reverse KL is the default distillation objective for LoCD
and the on-policy context-distillation baselines.
In single-step reasoning and coding tasks, a correction is limited to the length of the attributed original span. In multi-step agent tasks, the corrected step is used only for localized distillation; it is not executed in the environment and no continuation is collected from that correction.
A representative localization and attribution prompt is provided in
Appendix~\ref{app:attribution_prompts}.

\subsection{Evaluation Protocols}
\label{app:evaluation_protocols}

For reasoning and coding, \emph{avg@$K$} denotes the average fraction of correct responses across $K$ independently sampled generations per problem. For LiveCodeBench, the corresponding score is also reported as a pass@1 estimate based on 12 samples. All mathematical-reasoning and code-generation evaluations use a sampling temperature of $0.7$. Table~\ref{tab:appendix_reasoning_coding_eval} summarizes the dataset sizes and decoding limits.

\begin{table}[t]
\centering
\caption{Evaluation protocols for mathematical reasoning and code generation. The maximum number of new tokens is specified per sampled response.}
\label{tab:appendix_reasoning_coding_eval}
\small
\setlength{\tabcolsep}{5pt}
\renewcommand{\arraystretch}{1.12}
\begin{tabular}{lccc}
\toprule
\textbf{Benchmark} & \textbf{Problems} & \textbf{Metric} & \textbf{Max. new tokens} \\
\midrule
AIME24 & 30 & avg@12 & 38912 \\
AIME25 & 30 & avg@12 & 38912 \\
HMMT25-Feb & 30 & avg@12 & 38912 \\
HMMT25-Nov & 30 & avg@12 & 38912 \\
AMO & 50 & avg@12 & 38912 \\
MATH-500 & 500 & avg@1 & 38912 \\
OlympiadBench & 674 & avg@1 & 38912 \\
Minerva & 272 & avg@1 & 38912 \\
\midrule
CF-422 & 422 & avg@12 & 16384 \\
LiveCodeBench v6 & 175 & avg@12 & 16384 \\
\bottomrule
\end{tabular}
\end{table}

\paragraph{Interactive agents.}
We report task success rates using avg@1 on the complete 500-task WebShop test set and avg@3 on ALFWorld. Each WebShop episode is limited to 10 interaction steps, with up to 512 generated tokens per step during training and 256 at evaluation time; the interaction history is retained while earlier page observations are compressed. Each ALFWorld episode is limited to 50 steps, with up to 512 generated tokens per step in both training and evaluation; the model conditions on the two most recent observations. For ALFWorld, the main text reports performance on the 134-task \texttt{valid\_unseen} split.

\paragraph{Software-engineering agents.}
We report avg@4 for SWE-Gym, averaging success over four independent Mini-SWE-agent rollouts per instance on both the in-domain and out-of-distribution evaluation sets. On SWE-bench Verified, we report pass@1 over the full official 500-instance benchmark. We limit Mini-SWE-agent to 80 interaction steps per issue in both evaluations.

\paragraph{Model selection and reporting.}
For math reasoning and code generation, we evaluate all saved
checkpoints and report the best observed performance.
For WebShop, ALFWorld, and software-engineering tasks,
where evaluation is substantially more expensive due to
multi-step agent interactions, we instead select the
best-performing checkpoint on a held-out validation set
for final evaluation.
Specifically, WebShop and ALFWorld use 150-task subsets
of their development splits for checkpoint selection.
Results for math reasoning and code generation are reported
as mean $\pm$ standard deviation over three independent
training runs, while agent results are obtained from
a single training run per method.

\section{Pass@K Results.}
\label{app:passk}
\begin{figure}[h]
\centering
\includegraphics[width=\linewidth]{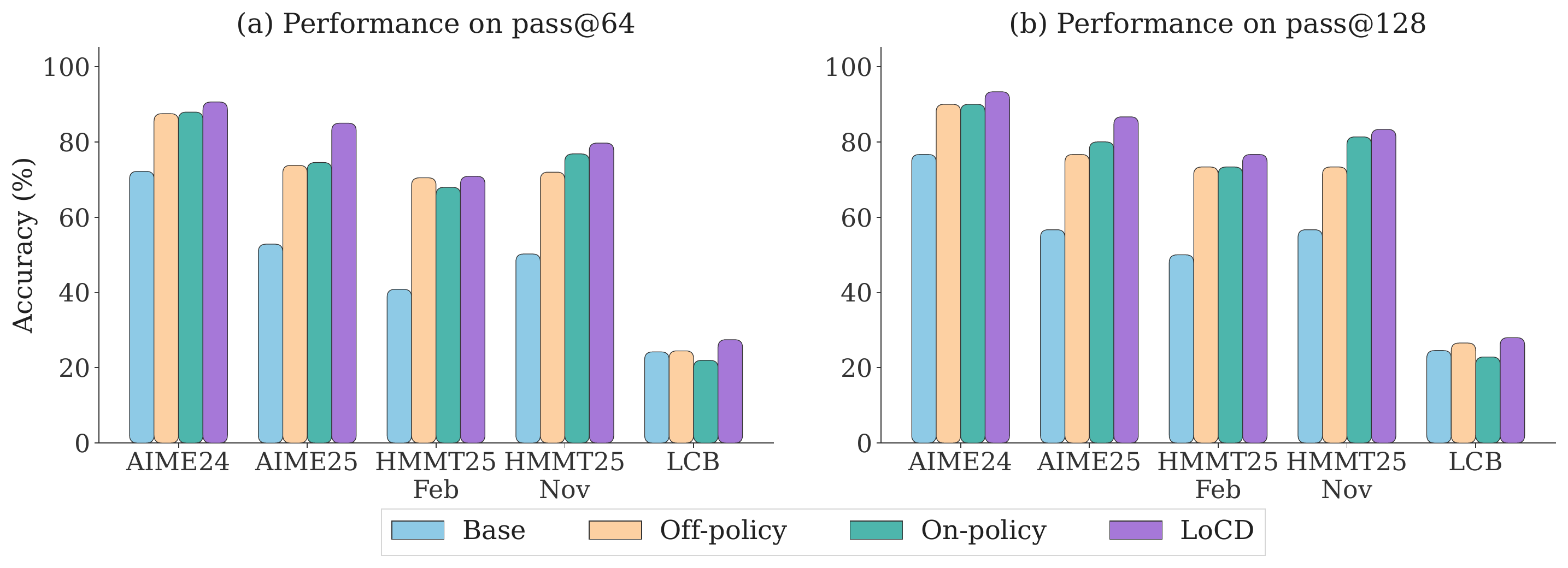}
\caption{Pass@K results on math and code: pass@64 (left) and pass@128 (right).}
\label{fig:passk}
\end{figure}

Figure~\ref{fig:passk} evaluates whether the gains of LoCD persist under larger
sampling budgets.
LoCD achieves the highest pass@64 and pass@128 across all five mathematics and
code benchmarks, consistently outperforming both off-policy and on-policy distillation.
The advantage remains visible as the number of samples increases, indicating
that LoCD improves not only single-sample accuracy but also the likelihood of
producing a successful solution under repeated sampling.

\section{Parameter-Update Geometry}
\label{app:geometry}

We provide additional details for the parameter-update analysis in
Section~\ref{sec:training_dynamics}.
Following the diagnostics of \citet{shen2026geometry}, we examine how different
context-distillation objectives modify the pretrained parameters over training.
We consider both the \emph{scale and support} of parameter changes and their
effect on the singular structure of the pretrained weights.
In addition to math reasoning reported in the main text, we conduct the
same analysis on code generation and ALFWorld.

\subsection{Analysis Protocol}
\label{app:geometry_protocol}

All analyses use Qwen3-8B and measure parameter changes relative to the shared
initial checkpoint.
For each weight matrix $W$, we denote its initialization by $W_0$ and its value
at training step $t$ by $W_t$.
We analyze checkpoints at
$t\in\{20,40,60,80,100\}$.
For ALFWorld, we use the same 100-step window to enable direct comparison with the reasoning and coding runs.

To cover different model depths while keeping the analysis tractable, we
analyze the query, key, value, and output projection matrices from every fourth
transformer layer,
$i\in\{0,4,8,\ldots,32\}$,
yielding 36 matrices in total.
Each diagnostic is computed independently for every matrix and then averaged
equally across the 36 matrices; thus, larger matrices do not receive additional
weight.

Stored weights are converted to FP32 before computing parameter differences.
Singular-value decompositions are computed in FP64 for numerical stability.
The same matrices, checkpoints, and aggregation procedure are used for all
methods and tasks.

\subsection{Metric Definitions}
\label{app:geometry_metrics}

We use four complementary diagnostics.
Update magnitude and visible update sparsity characterize how much and where
the parameters change, while principal angle and normalized spectral shift
characterize changes in the singular geometry of the pretrained weights.

\paragraph{Cumulative update magnitude.}
For each matrix, we measure its displacement from initialization as
\begin{equation}
    \Delta W_t = W_t-W_0,
    \qquad
    d_t(W)=\|\Delta W_t\|_F.
\end{equation}
This quantity measures the net displacement at step $t$, rather than the sum of
successive update norms.
We do not normalize it by matrix size or by $\|W_0\|_F$.
The reported update scale is the mean of $d_t(W)$ over the analyzed matrices.

\paragraph{Visible update sparsity.}
To distinguish substantial parameter movement from small numerical changes, we
mark an entry as visibly changed when
\begin{equation}
    M_{t,ij}
    =
    \mathbb{I}\!\left[
    |W_{t,ij}-W_{0,ij}|
    >
    \eta\max\!\left(|W_{0,ij}|,|W_{t,ij}|\right)
    \right],
    \qquad
    \eta=10^{-3}.
\end{equation}
For a matrix containing $n_W$ entries, we define
\begin{equation}
    \mathrm{Sp}_t(W)
    =
    1-\frac{1}{n_W}\sum_{i,j}M_{t,ij}.
\end{equation}
Higher values therefore indicate that a larger fraction of coordinates remain
within the relative-change threshold.
Equivalently, $1-\mathrm{Sp}_t(W)$ is the visibly changed-coordinate fraction
reported in our analysis.

This diagnostic follows the relative-change criterion used in prior geometric
analyses of distillation, but our implementation applies the criterion directly
to the stored weights after FP32 conversion and does not perform an additional
bf16 round trip.
We therefore interpret it as \emph{visible update sparsity}, rather than as an
indicator of whether a parameter received gradients.

\paragraph{Principal-subspace rotation.}
We next measure how much training rotates the dominant singular subspace.
Let
\begin{equation}
    W_0 = U_0\Sigma_0V_0^\top,
    \qquad
    W_t = U_t\Sigma_tV_t^\top.
\end{equation}
We retain the leading $k=512$ left singular vectors,
$U_{0,k}$ and $U_{t,k}$.
Their principal angles are
\begin{equation}
    \theta_{t,i}
    =
    \arccos
    \left(
    \sigma_i\!\left(U_{0,k}^{\top}U_{t,k}\right)
    \right),
    \qquad i=1,\ldots,k.
\end{equation}
The singular values of the overlap matrix are clipped to $[0,1]$ before
applying $\arccos$.
For each weight matrix, we use the median principal angle in degrees and then
average this statistic across matrices.
Smaller values indicate less rotation of the dominant left singular subspace.
This metric measures subspace orientation and should not be interpreted as the
rank of the parameter update.

\paragraph{Normalized spectral shift.}
Principal angles characterize changes in singular vectors; we additionally
measure changes in the singular values themselves.
Let $\sigma(W)$ denote the complete singular-value spectrum of $W$ in descending
order.
We compute
\begin{equation}
    \mathrm{NSS}_t(W)
    =
    \frac{
        \|\sigma(W_t)-\sigma(W_0)\|_2
    }{
        \|\sigma(W_0)\|_2
    }.
\end{equation}
Unlike the principal-angle diagnostic, NSS uses the full singular spectrum.
Lower NSS indicates a smaller relative change in the singular values of the
pretrained weight matrix.
The reported NSS is averaged equally across the analyzed matrices.

\begin{figure}[t]
\centering

\begin{subfigure}{0.9\textwidth}
    \centering
    \includegraphics[width=\linewidth]{figs/geo_main_math.pdf}
    \caption{Math reasoning.}
    \label{fig:geometry_math_appendix}
\end{subfigure}

\vspace{0.6em}

\begin{subfigure}{0.9\textwidth}
    \centering
    \includegraphics[width=\linewidth]{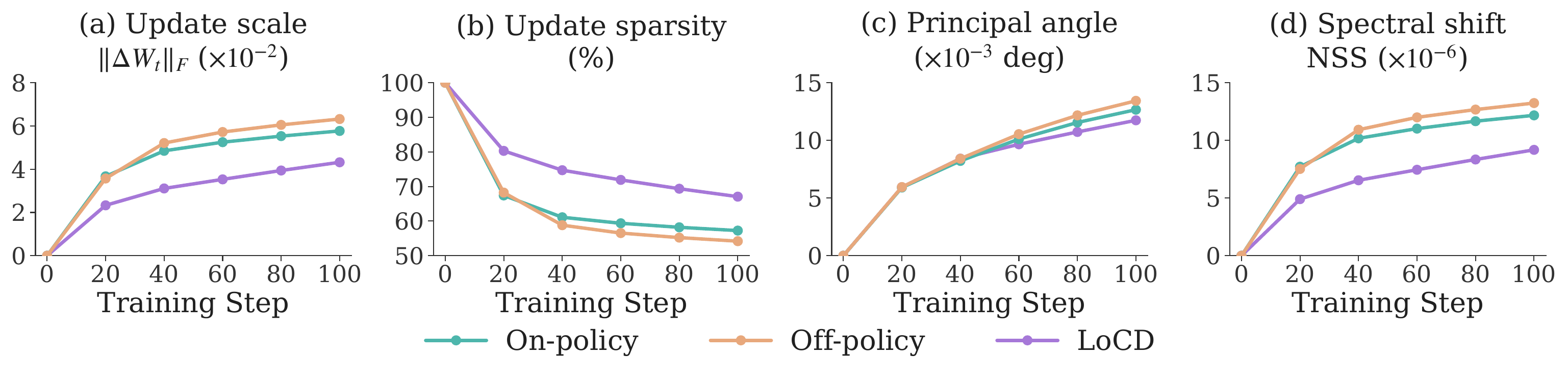}
    \caption{Code generation.}
    \label{fig:geometry_code}
\end{subfigure}

\vspace{0.6em}

\begin{subfigure}{0.9\textwidth}
    \centering
    \includegraphics[width=\linewidth]{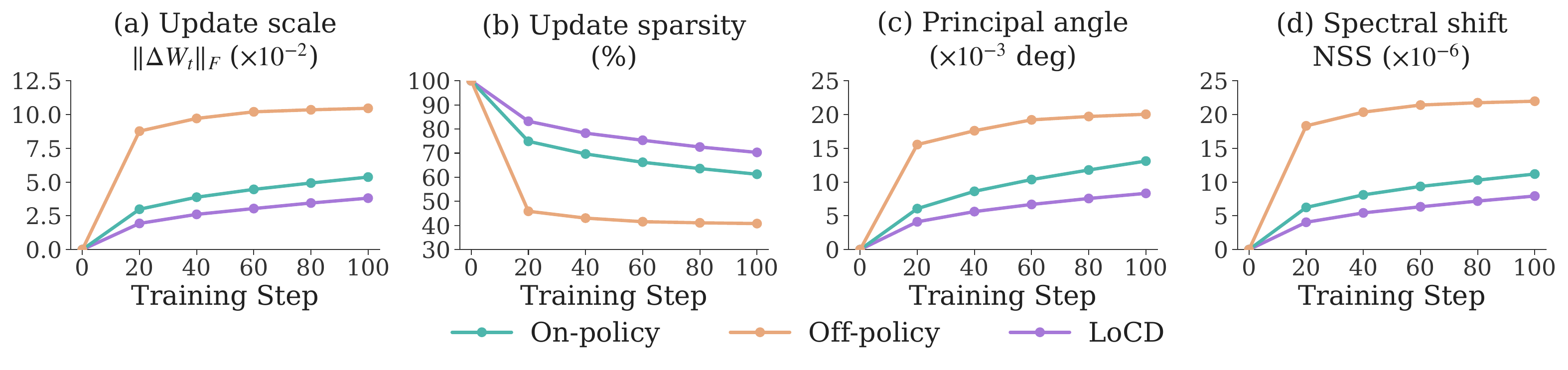}
    \caption{ALFWorld.}
    \label{fig:geometry_alfworld}
\end{subfigure}

\caption{
Parameter-update geometry across three datasets.
}
\label{fig:geometry_appendix_all}
\end{figure}

\subsection{Results Across Tasks}
\label{app:geometry_results}

Across the three settings, the most consistent distinction appears in update
magnitude and visible update sparsity: LoCD stays closer to initialization and
changes fewer coordinates than either on-policy or off-policy context
distillation.
The spectral diagnostics show a similar but somewhat weaker pattern.

\paragraph{Math reasoning.}
Figure~\ref{fig:geometry_math_appendix} reproduces the main-text analysis.
LoCD exhibits smaller cumulative displacement throughout training.
At step 100, its update magnitude is approximately 28--31\% smaller than those
of the on-policy and off-policy baselines.
The visibly changed-coordinate fraction is 32.66\%, compared with 44.60\% for
on-policy and 45.93\% for off-policy distillation.
LoCD also yields a lower NSS
($9.17\times10^{-6}$ versus $12.69\times10^{-6}$ and
$13.27\times10^{-6}$).
The principal-angle difference is comparatively modest
($0.01113^\circ$ versus $0.01182^\circ$ and $0.01192^\circ$),
indicating that the clearest separation lies in update magnitude and visible
support rather than a substantially different rotation of the dominant
subspace.


\paragraph{Code generation.}
Figure~\ref{fig:geometry_code} shows a closely matching pattern on code
generation.
At step 100, the mean update magnitude is $0.0432$ for LoCD, compared with
$0.0577$ for on-policy and $0.0632$ for off-policy distillation.
The corresponding visibly changed-coordinate fractions are 32.93\%, 42.74\%,
and 45.80\%, respectively.
LoCD also exhibits a smaller spectral shift.
Principal-angle differences are less pronounced and do not follow a strict
ordering at every checkpoint, again suggesting that the main distinction is in
the scale and support of the parameter update.

\paragraph{ALFWorld.}
The separation is even larger on ALFWorld
(Figure~\ref{fig:geometry_alfworld}), particularly relative to off-policy
distillation.
At step 100, the mean update magnitude is $0.0381$ for LoCD, $0.0537$ for
on-policy distillation, and $0.1046$ for off-policy distillation, making the
LoCD displacement 63.6\% smaller than the off-policy baseline.
The visibly changed-coordinate fractions are 29.70\%, 38.75\%, and 59.20\%,
respectively.
LoCD also exhibits smaller principal-subspace rotation and NSS throughout the
analyzed window.

\paragraph{Overall pattern.}
Across math reasoning, code generation, and ALFWorld, LoCD consistently
induces smaller cumulative displacement and a smaller fraction of visibly
changed coordinates in the analyzed attention projections.
Spectral changes are also generally smaller, while differences in dominant
subspace rotation are more modest on reasoning and coding.
Together with the task-performance results, these analyses show that LoCD
achieves stronger performance while modifying the pretrained parameters more
locally.

These diagnostics characterize the parameter changes observed in the analyzed attention projections.
They do not imply fewer trainable parameters, lower optimization cost, low-rank updates, or that smaller parameter changes are themselves the cause of LoCD's performance gains.

\section{Localization and Attribution Prompts}
\label{app:attribution_prompts}

We employ task-specific prompts for localization and
attribution across all evaluated tasks.
Here, we present mathematical reasoning as a representative
example to illustrate our prompt design.
Given a problem, the student's attempt, and reference
feedback, the model identifies critical reasoning errors
and generates corresponding problem-specific corrective hints.
The following template summarizes the key instructions
used in this process.

\begin{locdprompt}{Localization and Hint Generation in Math Tasks}
# Problem
{question}

# Student Attempt
{response}

# Reference Information (Privileged)
{feedback_block}

# Instructions

You are tasked with identifying critical reasoning errors in the student's attempt and providing targeted guidance for correcting them.

1. Error Identification
   Carefully examine the student attempt against the reference information, paying particular attention to the correctness feedback. If the final answer is incorrect, identify the earliest verifiable reasoning error. Return no attribution only if no specific error can be reliably identified.

2. Critical Span Localization
   Identify at most {max_states} substantive errors or clearly suboptimal steps that affect the solution. Prioritize the earliest error and ignore issues related only to wording, formatting, or presentation.

   For each identified error, copy one complete sentence verbatim from the student attempt to mark the correction point. The selected sentence should be concise (preferably within 200 characters) and must exactly match the original text. Do not select a fragment within a sentence, mathematical expression, or LaTeX block.

3. Corrective Hint Generation
   For each identified error, generate a concrete, process-level hint specific to the current problem. The hint should:
   - Describe the relevant mathematical situation, explicitly referring to the quantities, equations, or relationships involved.
   - Specify an actionable direction for correcting the reasoning.
   - When helpful, explain which incorrect approach should be avoided and why.

   Avoid vague suggestions or generic mathematical advice. Do not reveal the final answer or reproduce the reference solution.

4. Correction Context
   The selected erroneous sentence and all subsequent content will be hidden during correction. The hint must therefore be self-contained and understandable from the preceding student prefix. Do not refer to deleted content using expressions such as "the above step" or "this calculation".

# Output Format
Return strict JSON without additional text:

{
  "attributions": [
    {
      "span": "<verbatim sentence marking the error>",
      "hint": "<concrete, problem-specific corrective hint>"
    }
  ]
}
\end{locdprompt}

The returned \texttt{span} serves as a textual anchor
for locating the correction point in the student trajectory,
while the corresponding \texttt{hint} provides privileged
guidance for localized teacher generation.
As described in Section~\ref{sec:method},
distillation is restricted to the corrected segment
rather than the entire remaining trajectory.

\end{document}